\pdfoutput=1

\documentclass[11pt]{article}

\usepackage{multirow}
\usepackage{graphicx}
\usepackage{subfigure}

\usepackage{algorithm}
\usepackage {algorithmicx, algpseudocode}

\usepackage{authblk}
\author{Seok-Ung Choi}
\author{Joonghyuk Hahn}
\author{Yo-Sub Han}

\affil{School of Computing, Yonsei University, Seoul, Republic of Korea}

\usepackage[utf8]{inputenc} 
\usepackage[T1]{fontenc} 

\usepackage{fullpage}

\usepackage{url}            
\usepackage{booktabs}       
\usepackage{amsfonts}       
\usepackage{nicefrac}       
\usepackage{microtype}      
\usepackage[dvipsnames]{xcolor}

\usepackage{natbib} 
\usepackage{enumitem}
\usepackage{amsthm}
\usepackage{amssymb}
\usepackage{amsmath}
\usepackage{mathrsfs}
\usepackage{thmtools}
\usepackage{hyperref}
\hypersetup{
    pdftoolbar=true,        
    pdfmenubar=true,        
    pdffitwindow=false,     
    pdfstartview={FitH},    
    pdfnewwindow=true,      
    colorlinks=true,       
    linkcolor=blue,          
    citecolor=ForestGreen,        
    filecolor=magenta,      
    urlcolor=red,           
    breaklinks=false,
}

\usepackage{cleveref}
\usepackage{thmtools}
\usepackage{thm-restate}

\numberwithin{equation}{section}

\date{}

\title{URECA: The Chain of Two Minimum Set Cover Problems exists behind Adaptation to Shifts in Semantic Code Search}

\newtheorem{definition}{Definition}[section] 
\newtheorem{theorem}{Theorem}[section] 
\newtheorem*{theorem*}{Theorem} 
\newtheorem{corollary}{Corollary}[section] 

\newtheorem{lemma}{Lemma}[section]

\begin{document}

\maketitle

\begin{abstract} 
Adaptation is to make model learn the patterns shifted from the training distribution. 
In general, this adaptation is formulated as the minimum entropy problem. 
However, the minimum entropy problem has inherent limitation---shifted initialization cascade phenomenon. 
We extend the relationship between the minimum entropy problem and the minimum set cover problem via Lebesgue integral. 
This extension reveals that internal mechanism of the minimum entropy problem ignores the relationship between disentangled representations,
which leads to shifted initialization cascade.
From the analysis, we introduce a new clustering algorithm, Union-find based Recursive Clustering Algorithm~(URECA).
URECA is an efficient clustering algorithm for the leverage of the relationships between disentangled representations.
The update rule of URECA depends on Thresholdly-Updatable Stationary Assumption to dynamics as a released version of Stationary Assumption.
This assumption helps URECA to transport disentangled representations with no errors based on the relationships between disentangled representations.
URECA also utilize simulation trick to efficiently cluster disentangled representations. 
The wide range of evaluations show that URECA achieves consistent performance gains for the few-shot adaptation 
to diverse types of shifts along with advancement to State-of-The-Art performance in CoSQA 
in the scenario of query shift. 
\end{abstract}

\newpage
\tableofcontents
\newpage

\section{Introduction}
\label{sec:intro}

Code search is a task to retrieve code snippets from the given query whose intent is to find implementations of specific functionality. 
This task is important for both human developers' productivity and LLM's hallucination with RAG 
(\cite{BrownBNMJPA20, LewisEAFVNHMWT20, LeiWWZHQWXBT25}). 
Although the scale-up of programming language models boosts performance of this task, 
the models still are vulnerable to diverse types of shifts (\cite{ArakelyanDMR23}), 
which comes from the evolutions that human developers continuously write new codes for debug or performance improvements.
Adaptation to these evolutions relies on generalization, 
a key mechanism of human intelligence that also drives the application of neural networks across wide range of fields.

We can utilize supervised signals as catalysts for the generalization. 
In spite of the effectiveness, 
only insufficient supervised signals are available as guidances to generalization
due to severity of shifts.
Therefore, it is necessary to analyze adaptation itself for effective utilization of these supervised signals.
Adaptation to shifts is formulated as minimum entropy problem in code search.
However, minimum entropy problem cannot help shifted initialization lead to bad solution (shifted initialization cascade phenomenon).
(\cite{OriRYM24}) demonstrates that this phenomenon occurs in entropy minimization~(\cite
{Yvew04}) 
which is the unsupervised instantiation of minimum entropy problem.

\begin{figure}[h]
\centering
\includegraphics[width=1.\columnwidth]{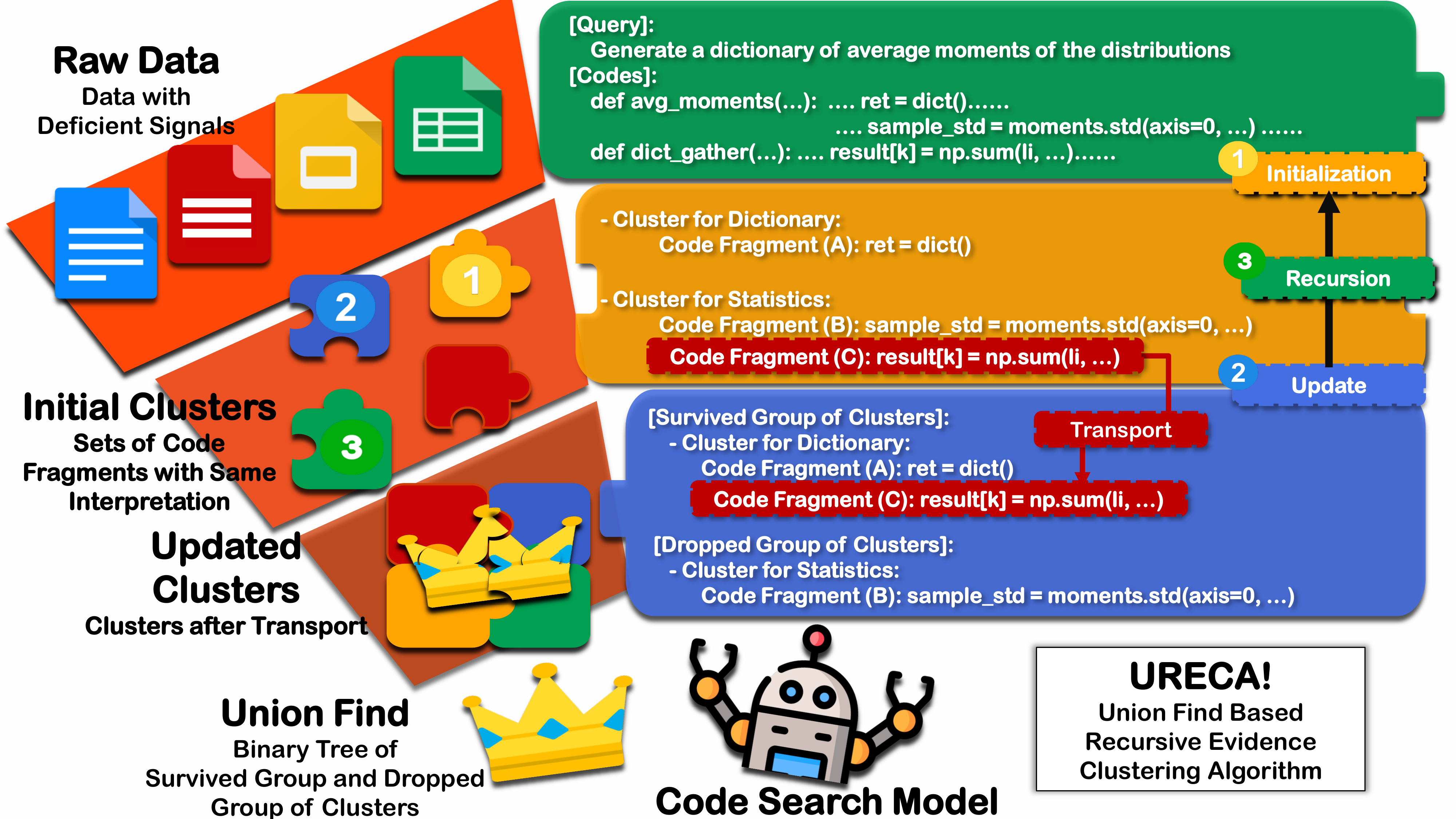} 
\caption{
Hypothetical Example of Real World Scenario in Figure 2.
URECA simulates Real World Scenario with simulation trick.
}
\label{fig1}
\end{figure}

We break down the minimum entropy problem in the lens of Lebesgue integral 
to analyze the mechanics of this shifted initialization cascade.
In the process of decomposition, we extend the connection of minimum entropy problem 
to minimum set cover problem towards the unknowns. 
As a result, we find out that minimum entropy problem hides the chain of two minimum set cover problems equipped with greedy algorithm.
This allows us to discover internal mechanism of minimum entropy problem, 
which makes lumps of disentangled representations with little regard to relationships between them.
(\cite{LeeJLPSHY24}) supports this result with analysis 
that entropy minimization neglects the influence of these disentangled representations, 
which leads to unreliable predictions in biased scenarios. 
In addition, (\cite{WilesGSRKDC22}) assists that it is important to understand the disentangled representations for adaptation to distribution shift.
In succession to these insights, 
we focus on the ignorance of minimum entropy problem to the relationships between disentangled representations
and demonstrate that this ignorance leads to shifted initialization cascade in Section 2. 

From this analysis,  we introduce a new clustering algorithm, URECA.
URECA clusters disentangled representations based on the relationships between them. 
URECA consists of three steps, initialization, update and recursion as depicted in Figure 1.
In the process of intialization, URECA fragmentizes samples and initializes clusters with those fragments.
(In this paper, evidence refers to fragment and logit is the weight of clusters of evidences. 
In addition, disentangled representation is the encoding for evidence / fragment.)
We apply the analyses of Theorem 2.1 and 2.2 to the scenario of training neral network.
Then, theoretical relevance of logits and conditional probabilities reveals that the forward propagation and dot product do 
the fragmentization of samples.
Then, minimum entropy problem initializes the cluster based on these fragments.
URECA utilizes this mechanism of neural networks to initialize clusters of fragments. 

In update step, URECA efficiently updates the clusters with transport of fragments from source clusters to target clusters.
Simulation trick makes it possible for URECA to simulate this process with simple calculations of weights of evidences 
for clusters~(\cite{Greiff00}). 
In this paper, we use logits as these weights of evidences for this simulation trick due to the excessive regularization of probabilities.
However, the naive use of logits triggers errors in the process of transport.
We treat this problem by introducing Thresholdly-Updatable Stationary Assumption 
for dynamics.
This assumption makes dynamics to more correctly reflect on the relationships between disentangled representations
by filtering out irrelevant disentangled representations. 
The accurate reflection makes the transported logits as unbiased estimator for the logits of transported probabilities,
which clears out the errors from naive replacement about probabilities.
For the next, we determine the codes in reference to the clusters which are constructed by URECA based on 
the relationships between disentangled representations. 
Finally, we introduce an auxiliary loss about the relationships between codes as media to leverage the relationships between disentangled 
representations, since they are estimated based on the relationships between disentangled representations. 

Our extensive evaluations demonstrate that URECA is consistently superior to baselines for few shot adaptation to diverse shifts. 
Few shot adaptation to shifts is a realistic setting since the harshness of shifts makes harder 
to collect sufficient data in limited time.
In addition, we architect experiments to enable comparative analysis 
through coordination of datasets and baselines for diverse types of shifts like task shift, query shift and 
code shift~(Appendix F.1 $\sim$ F.3).  
These intricate designs for experiments effectively highlight that URECA drives model to general patterns 
which reflect on the relationships between disentangled reprensentations.

Our contributions  are as follows: 
\begin{itemize}
    \item As long as we know, we are the first to prove that the chain of two minimum set cover problems 
          exists behind minimum entropy problem in terms of Lebesgue integral.
          We present analysis that this chain ignores relations between disentangled representations.
          We also prove by construction, for the first time, 
          that minimum entropy problem clusters disentangled representations. 
          
    \item For effective adaptation to diverse shifts,
          we propose new clustering algorithm URECA which reflects on the relationship 
          between disentangled representations.
          URECA simulates the process of clustering based on simulation trick and 
          overrides Stationary Assumption with Threshold-Updatable Stationary 
          Assumption for dynamics to correctly reflect on
          the relationships between disentangled representations .           
    \item We carefully design the structure of experiments for comparative analysis across diverse 
          types of shifts. 
          This architecture demonstrates that URECA effectively adapts to diverse types of shifts
          like task shift, query shift and code shift. 
          Especially, URECA achieves State-of-The-Art performance in CoSQA 
          along with consistent performance gains across various programming languages 
          and shifts in CSN. 
\end{itemize}

\section{Problem Analysis}
At first, we analyze the minimum entropy problem in the lens of Lebesgue integral (Theorem 2.1).
After then, we show that model learns to cluster the elements by minimum entropy problem (Theorem 2.2).
These theorems drive us to notice that the ignorance of disentangled relationships leads to shifted initialization cascade. 

\begin{theorem}
\label{thm:min_entropy_min_setcover}
Minimum entropy problem is dual to the combination of chains of two minimum set cover problems with greedy algorithm for disentangled representations. 
\end{theorem}
\begin{align}
\label{eq:thm_min_entropy_min_setcover}
\ln {\frac{1}{p(E_\alpha)}} &=\int p(E_\alpha)d{\frac{1}{p(E_\alpha)}}\\
                          &=\sup\{\sum_{n=1}^{\lceil{\frac{1}{p(E_\alpha)}}\rceil}p(E_\alpha)\}\\  
\begin{split}
                          &=\sup\{\sum_{n=1}^{\lceil{\frac{1}{p(E_\alpha)}}\rceil}\min_{u \subset U}\int_e s(e)d\chi(e \in u)\}\\
                          &(s.t.\ p(E_\alpha) \leq p(u))
\end{split}
\end{align}
As we mentioned, we examine minimum entropy problem in the lens of Lebesgue integral.
The proof of Theorem 2.1 is provided in Appendix D.1.
$u$ is the subset of $U$ which is the universe of disentangled representations $e$ for the predecessor minimum set cover problem.
The predecessor minimum set cover problem is the first problem of the chain of two minimum set cover problems
(We call the last problem of the chain as successor minimum set cover problem).
$\chi(e)$ is the characteristic function which describes whether $e$ is the element of $u$ (Appendix B).
$s(e)$ is the simple function which measures the size of $e$.

The analysis starts from equation (2.1) which describes the self-information in terms of Lebesgue integral to probability.
We can rewrite the integral form of equation (2.1) to supremum form of equation (2.2) based on the definition of Lebesgue integral 
for non-negative measure (Appendix B).
With properties of probability as Lebesgue measure, this supremum form hints that self information becomes the tight upper bound 
for the cost of solution to the chain of minimum set cover problems equation (2.3). 
This self information linearly combines to form the entropy.
Linear combinations do not change the direction of inequality in equation (36) and (37).
As a result, minimum entropy problem is dual form of the combination of chains of two minimum set cover problems by weak duality theorem.

Theorem 2.2 relates minimum entropy problem to clustering based on the previous analysis.
We provide the details about proof and examples in Appendix D.2 and Appendix D.3.
\begin{theorem}
\label{thm:min_entropy_cluster}
Minimum entropy problem is equivalent to the problem which clusters the disentangled representations 
to minimize the expected cost of clustering given the probabilities for each event.
\end{theorem}
According to proof of Theorem 2.2, 
the chain of minimum set cover problems constructs the set of clusters for each event.
At first, the predecessor problem gathers subsets of disentangled representations with greedy algorithm and 
initializes the universe of successor problem with those subsets like $\{\{00,10\},\{01, 11\}\}$ from 1st MSC in Figure 5.
Then, the successor also constructs the solution with greedy algorithm like $\{\{\{00, 10\}\},\{\{01, 11\}\}\}$ from 2nd MSC in Figure 5.
This, in turn, initializes the combination process for solutions of each chain of problems.
For all events, minimum entropy problem does the same thing for different events.
In the process of combination, computation selects only one element in the solution of chained problems for each event
like $\{\{00, 10\}\}$ for event $a$ in Figure 8.
After then, it combines them to new family of subsets like $\{\{\{00, 10\}\}, \{\{01\}\}, \{\{11\}\}\}$, 
which can be interpreted as clusters of disentangled representation by the definition of clustering (Definition D.1 in Appendix D.2).

From Theorem 2.1 and Theorem 2.2, we can construct the minium entropy problems 
with the combination of the minimum set cover problems.
The construction unveils the internal mechanism of minimum entropy problem, 
which reveals the limitation of minimum entropy problem in terms of clustering.
In the process of clustering, predecessor makes a series of greedy decisions for the construction of solution according to the measure 
that simply sums all probabilities of elements in specific subset.
This simple summation of elements means that the decision criteria does not consider the relationships between elements 
(disentangled representations) but only considers about the size of each element itself 
unlike general clustering which considers about the distance between elements. 

The analysis also holds true for InfoNCE~(\cite{DenYO18}) since it considers the only relationships between given samples
which are mixtures of disentangled representations.
The coarse-grained relationships limit for model to capture the fine-grained relationships between disentangled representations.
This becomes apparent in the situation that fragments in same sample can sometimes be less relevant to each other than fragments of different sample.
For example, Code Fragment (C) in Figure 1 is less relevant to the Code Fragment (A) of same function than Code Fragment (C) of different function.  
Figure 3 demonstrates that shifted initialization cascade comes from the ignorance of relationships between disentangled representations 
as encoding of these fragments.

\section{Union-find based Recursive Evidence Clustering Algorithm (URECA)}

\begin{figure*}[t]
\centering
\includegraphics[width=1.\textwidth]{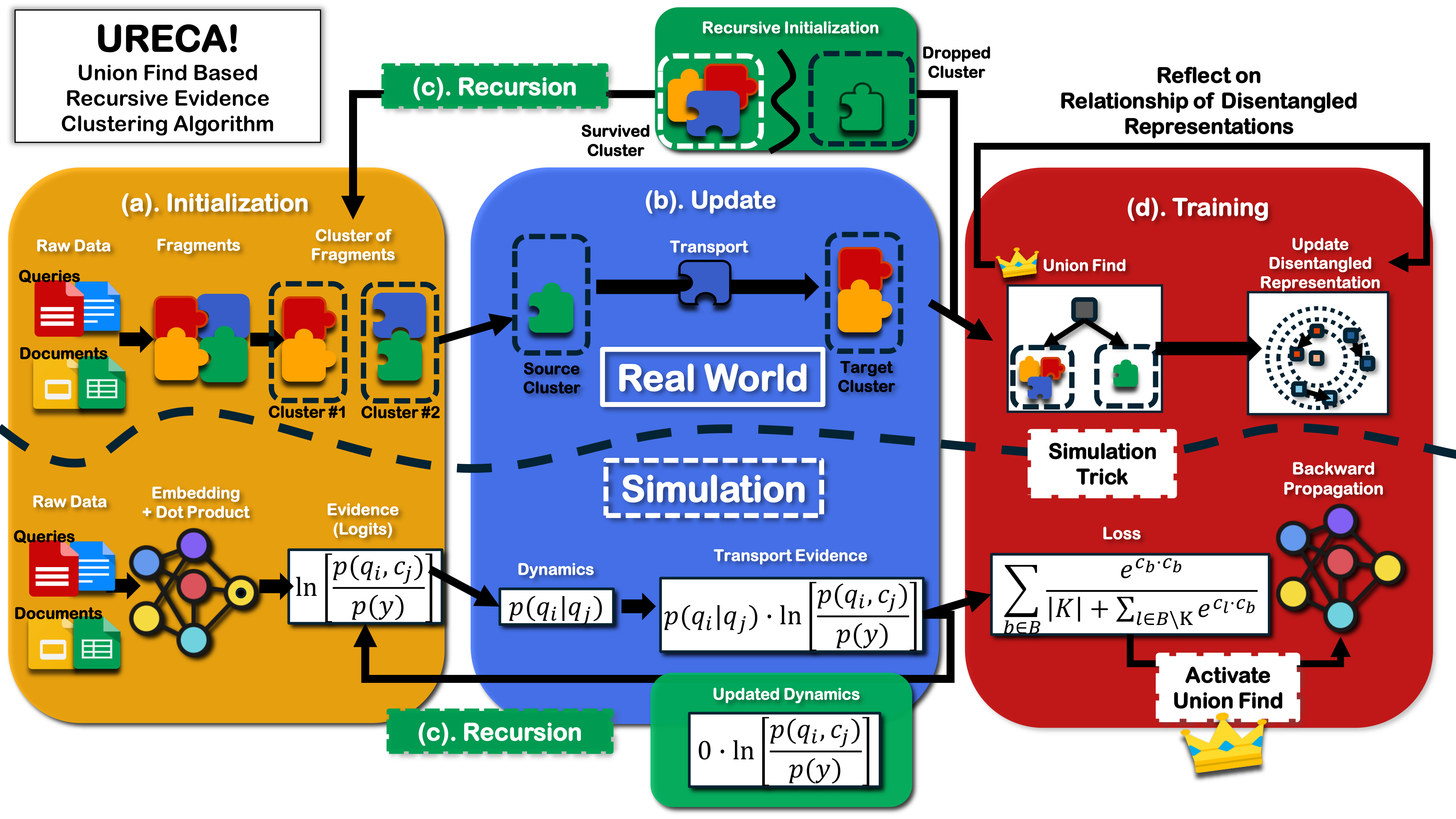} 
\caption{
Overview of URECA
}
\label{fig2}
\end{figure*}

URECA is a new algorithm to cluster disentangled representations based on the relationships between disentangled representations.
URECA implements the iterative steps of initialization, update and recursion in other clustering algorithms (k-means etc.) 
with the combination of calculations for numeric values based on simulation trick. 
Simulation trick is the trick to replace specific task to simple numeric computations. 
For example, a simulation implements moving a box only by changing its position 
from (1, 2, 3) to (4, 5, 6) with the addition of 3 for each element of (1, 2, 3).
Figure 2 illustrates the algorithm of URECA in terms of Real World Scenario and Simulation Scenario.  
We also provide pseudo code of URECA in Appendix C and details of analyses in Appendix E.

\subsection{Initializatoin Step}
At first, URECA calculates weight of evidence for initial cluster (logits) like equation (4).
$(q_i,c_i)$ is the given pair of (query, code) and $f$ is neural network.
\begin{align}
\label{initial_evidence_cluster}
\begin{split}
    evid_0(C^0_{i,j})&=\ln{ \frac{p(q_i,c_j)}{p(y)}}\\
                     &= f(q_i) \cdot f(c_j)
\end{split}
\end{align}
In the context of code search, the estimation of weight of clusters leads to Fragment Initialization and Cluster Initialization.
We can justify this statement since the estimation of logits leads to estimation of conditional probability and entropy, 
which respectively correspond to Fragment Initialization and Cluster Initialization.

\textbf{Fragment Initialization: }
We can apply the analysis of Theorem 2.1 and Theorem 2.2 to conditional probability as output of neural network
in the scenario of training neural network with minimum entropy problem.
Neural network calculates conditional probability based on logits and softmax.
Conditioning random variable in this conditional probability prerequisites the inverse image as subset of universe of fragments. 
The conditional probability configures this subset of universe of fragments as universe of the predecessor minimum set cover problem.
In other words, forward propagation and dot product adopt the only query-relevant fragments in code search
and do fragment initialization with them.
This conditional probability also leads to configuration of candidates and criteria 
for greedy decision to the chain of minimum set cover problems

\textbf{Cluster Initialization: }
Then, the calculation of entropy clusters the query-relevant fragments like Figure 5, 6, 7 and 8. 
The chain of two minimum set cover problems operate to cluster disentangled representations for each event.
Then, the calculation of weighted self-information($p\ln{\frac{1}{p}}$) selects only one cluster 
in the output of the chain for each event.
This statement can be justified since all elements of solution have probability equal to each other.  
At last, the cluster initialization completes by gathering the selected clusters across all events.

\subsection{Update Step}
Update step consists of dynamics estimation and transport computation.
URECA esitmates dynamics of disentangled representations with attention scores of queries. 
After then, URECA estimtates the transported parts of logits through linear combination of logits with these dynamics.
Then, it subtracts the parts of logits from logits of dropped cluster and adds the value to logits of survived cluster.
The subtraction and addition simulate to transport evidences/fragments from dropped cluster to survived cluster.

Simulation trick enables URECA to simulate the transport of fragments with numeric calculations.
The following properties of Lebesgue measure enable simulation trick to operate with no errors (equation (3.2)).
\begin{align}
\label{lebesgue_property_union_diff}
\begin{split}
\mu(A \cup B) &= \mu(A)+\mu(B) - \mu(A \cap B)\\ 
\mu(A-&B)=\mu(A)-\mu(A \cap B)
\end{split}
\end{align}
These properties naturally connects the numeric calculations for Lebesgue measure to transport elements between measurable set $A$ and $B$.
We split the transport into two sub-units, separation from $C_s$ (subtraction) and clump into $C_t$ (addition).
To begin with, we can formalize the separation to separate semantic fragment $R_{s,t}$ from $C_s$ as follows.
\begin{align}
\label{lebesgue_property_diff_fragment}
\begin{split}
\mu(C_s-R_{s,t})  &= \mu(C_s)-\mu(C_s \cap R_{s,t})\\
                  &=\mu(C_s)-\mu(R_{s,t})
\end{split}
\end{align}
Then, we can also derive the following formula for addition to clump $R_{s,t}$ into $C_t$.
\begin{align}
\label{lebesgue_property_union_fragment}
\begin{split}
\mu(C_t \cup R_{s,t})=\mu(C_t)+\mu(R_{s,t})-\mu(C_t \cap R_{s,t})
\end{split}
\end{align}
The estimation of $\mu(R_{s,t})$ enables to subtract the estimation from $\mu(C_s)$ and add it to $\mu(C_t)$, 
which simulates the transport subset of semantic fragments($R_{s,t}$) from $C_s$ to $C_t$
(In this case, $\mu(C_t \cap R_{s,t}) $ should be zero by the assumption that specific fragment/evidence belongs 
to the only one cluster at specific time).
In other words, we can accurately implement the update of general clustering algorithm 
with the sequence of operations for Lebesgue measure.
Probability meets the basic properties in equation (3.2) since it is one of the Lebesgue measures.
Therefore, we can formulate the transport as follows with Bayesian theorem.
\begin{align}
\label{evidence_for_prob}
\begin{split}
evid(C^{t+1}_{i,i})&=E_{j\sim p(C^{t+1}_{i,i}|C^t_{j,j})}[p(C^t_{j,j})]
\end{split}
\end{align}
However, probability does not operate well for the training of neural network due to the regularized property in some settings, 
especially for few shot learning.
To avoid the excessive regularization, 
we replace the probabilities ($p(C^t_{j,j})$) to logits ($\ln{\frac{p(C^t_{j,j})}{p(y)}}$).
\begin{align}
\label{evidence_for_logits}
\begin{split}
evid(C^{t+1}_{i,i})=E_{j\sim p(C^{t+1}_{i,i}|C^t{j,j})}[\ln{\frac{p(C^t_{j,j})}{p(y)}}]
\end{split}
\end{align}
This naive replacement withdraws the theoretical guarantees about the accurate simulation to transport 
since logit is not the Lebesuge measure.
As a result of efforts to theoretical guarntees for this update rule (equation (3.6)), 
we specify the conditions for the expectation of logits to become unbiased estimator with Theorem 3.1 and Corollary 3.2 in the next sub-section.

In this update rule, the dynamics becomes $p(C^{t+1}_{i,i}|C^t_{j,j})$ since evidence is updated 
based on the conditional probability.
\begin{align}
\label{orig_dynamics}
\begin{split}
dyn(C^{t+1}_{i,i},C^t_{j,j})=p(C^{t+1}_{i,i}|C^t_{j,j})
\end{split}
\end{align}
We assume that the conditional probability as dynamics is stationary for each time step of update.
\begin{align}
\label{stationary_dynamics}
\begin{split}
dyn(C^{t+1}_{i,i},C^t_{j,j})&=p(C^{t+1}_{i,i}|C^t_{j,j})\\
                          &=p(C_{i,i}|C_{j,j})\ \ (\forall t \in N)
\end{split}
\end{align}
We also hypothesize that query $q_i$ is the centroid of cluster $C_{i,j}$ for all $j$.
As a result, we calculate the dynamics with attention scores across queries
since they represent the properties of corresponding clusters as centroids. 
\begin{align}
\label{dynamics_impl}
\begin{split}
p(C_{i,i}|C_{j,j})&=p(q_{i}|q_{j})\ \ (\forall t \in N)
\end{split}
\end{align}

For the next sub-section, we release the Stationary Assumption to Thresholdly Updatable Stationary Assumption. 
We model this stationary conditional probability with attention score for queries
since the softmax operation allows to estimate the conditional probability in the process of calculation for attention scores.
We provide the reason why we only focus on the query in Appendix E.1.

\subsection{Recursion Step} 
As mentioned earlier, we use logits instead of probabilities  
which impede training of neural network due to excessive regularization.
However, logits do not meet the properties of Lebesgue measure.
This unsatisfaction causes errors in the process of transport-based estimation for next time step’s logits.
Probability-based transport decides next time step's logits as follows. 
\begin{align}
\label{accurate_transport}
\begin{split}
\ln{p_{t+1}(C^{t+1}_{i,i}) \over p(y)}= \ln E_{j\sim p(C^{t+1}_{i,i}|C^t_{j,j})}[{p(C^t_{j,j})\over p(y)}]
\end{split}
\end{align}
In contrast, logit-based transport estimates next time step's logits as follows. 
\begin{align}
\label{estiamted_transport}
\begin{split}
evid(C^{t+1}_{i,i})=E_{j \sim p_{t+1}(C^{t+1}_{i,i}|C^t_{j,j})}[\ln{p_t(C^t_{j,j}) \over p(y)}]
\end{split}
\end{align}
According to Jensen’s Inequality, equation (3.10) and equation (3.11), the following inequality holds true 
between original transport and estimated transport.

\begin{align}
\label{estimation_error}
\begin{split}
\ln{p_{t+1}(C^{t+1}_{i,i}) \over p(y)}   \ge E_{j \sim p(C^{t+1}_{i,i}|C^t_{j,j})}[\ln{p_{t}(C^t_{j,j})\over p(y)}]
\end{split}
\end{align}
If we scrutinize the inequality, the estimated logit is smaller than original logit in general, 
which means there are positive errors for the estimation.
However, the error may disappear when all of the $p_t(C^t_{j,j})$s are equal to each other for all $j$s, 
since it satisfies the equality condition of Jensen’s inequality.
From this intuition, we derive Theorem 3.1 that describes the conditions for the uniform convergence of estimated logits to each other.
(UP-Limit is defined in Appendix D.4 as Definition D.2).
Uniform convergence assures that the error of estimation disappears as update proceeds.
The UP-Limit and Lipchitz continuity conditions for uniform convergence 
ensure that transported logits serves as unbiased estimator for the original logits.
We provide the proof of Theorem 3.1 in Appendix D.4. 
\begin{theorem}
\label{thm:uniform_convergence}
If the UP-Limit of $p_t(C^t_{j,j})$ converges to ${1 \over |J|}$ and 
$\ln {p_t(C^t_{j,j}) \over p(y)}$ is 
$\alpha\cdot\ln{{1 \over |J|} +\epsilon \over {1 \over |J|}-\epsilon}$-Lipschitz continuous 
($\forall \alpha \in [0,{{1 \over |J|} -\epsilon \over {1 \over |J|} +\epsilon})$ 
and $\forall \epsilon \in [0,\infty)$), 
then $\ln {p_t(C^t_{j,j}) \over p(y)}$ converges uniformly to each other as $t \rightarrow \infty$.
\end{theorem}
 
For the uniform convergence, we don’t have to care about the term whose $p_{t+1}(C^{t+1}_{i,i}|C^t_{j,j})$ 
becomes zero, since each term of RHS in equation (3.12) consists of $\ln {p_t(C^t_{j,j})\over p(y)}$s 
multiplied by $p_{t+1}(C^{t+1}_{i,i}|C^t_{j,j})$s. 
In this case, $p_t(C^t_{j,j})$ should converges to ${1 \over |K|}$ for uniform convergence like 
The definition of $K$ is provided with the proof of Corollary 3.1 in Appendix D.4
\begin{corollary}
\label{cor:restricted_uniform_convergence}
If $p_t(C^t_{j,j})$ converges to ${1 \over |K|}$ for UP-Limit and $\ln {p_t(C^t_{j,j}) \over p(y)}$  
is $\alpha\cdot\ln{{1 \over |J|} +\epsilon \over {1 \over |J|}-\epsilon}$-Lipschitz continuous 
($\forall \alpha \in [0,{{1 \over |J|} -\epsilon \over {1 \over |J|} +\epsilon})$ 
and $\forall \epsilon \in [0,\infty)$), then $\ln {p_t(C^t_{j,j}) \over p(y)}$ uniformly converges.
\end{corollary}
For the Corollary 3.2, $p_{t+1}(C^{t+1}_{i,i}|C^t_{j,j})=0$ implies that  $C^{t+1}_{i,i}$and $C^t_{j,j}$ 
are mutually exclusive (or disjoint) to each other.
Corollary 3.2 also assures that the mutual exclusion of $C^t_{i,i}$ and $C^{t}_{j,j}$.
The sole stationary assumption in previous sub-section cannot reflect on this property for uniform convergence of update.
Therefore, we introduce Threshold-Bounded Stationary Assumption which incorporates this property 
into the update rule.
This assumption means that dynamics are stationary until threshold $T$ 
but some $p_{t+1}(C^{t+1}_{i,i}|C^t_{j,j})$s reduce to zero once the threshold is crossed..

To implement this assumption, URECA estimates clusters mutually exclusive with the ground truth cluster 
and reduces the corresponding elements of dynamics to zero.
To estimate the disjoint clusters, 
we decide the sign of total divergence for clusters of all codes in terms of ground truth query.
Divergence is defined as the difference between intake and leak of elements for infinitesimal space. 
From this perspective, negative divergence means the leak of evidences to other clusters, 
which makes the cluster completely disjoint to other clusters as update progresses based on the current dynamics.
We provide the details about divergence and the analysis in Appendix E.2. 

In origin, total divergence of specific code $k$ in terms of query $i$ should consider 
about unit divergences for all code $j$s in equation (72) in Appendix E.2.
However, URECA narrows down the range of $j$ to ground truth code $i$ for computational efficiency.
In other words, URECA estimates the original total divergence as unit divergence of cluster $C^t_{i,k}$ with 
ground truth cluster $C^t_{i,i}$.
This unit divergence estimates the clusters disjoint to other clusters if unit divergence of $(q_i,d_k)$ is negative. 
This estimation results in the split of clusters into survived group (non-negative/possibly overlapped clusters) 
and dropped group (negative/disjoint clusters).
With the estimated groups, URECA clears out the elements of the dynamics that belong to the dropped group.
This process makes up of union find structure as binary tree, 
since URECA recursively iterates this process for each level which consists of survived node and dropped node 
which are disjoint to each other. 

URECA continues the recursion until transport is not possible anymore.
In terms of implementation, the transport becomes impossible when URECA cannot split the clusters into survived group 
and dropped group.
URECA cannot split the clusters when there is no negative divergence.
We naively assume that the no negative divergences come from the convergence of the joint probability distribution $p(C^t_{j,j})$ 
to uniform distribution as ${1 \over |K|}$, 
which is one of two conditions for uniform convergence of evidence transportation in the Corollary 3.2. 

However, $|K| > 1$ means that neural network does not know the difference between these clusters in $|K|$, 
since all $\ln {p_t(C^t_{j,j}) \over p(y)}$s are equal to each other for all $j$s.
This is the reason why we exclude the clusters for the calculation of loss ($L$ is the set of codes corresponding to dropped clusters).
\begin{align}
\label{URECA_loss}
\begin{split}
L_U= -{1 \over |B|}\sum_{b \in B}\ln{ exp(c_b \cdot c_b)\over |K| + \sum_{l \in L}exp(c_b\cdot c_l)}
\end{split}
\end{align}
While $L_U$ is derived from InfoNCE, it differs from the original InfoNCE in that it highlights the relationships between disentangled representations. 
It achieves this through the leverage of the estimated connections between codes, 
which align with the relationships between clusters estimated from URECA.
This $L_U$ is an auxiliary loss for $L_I$, the original InfoNCE loss between query and code,
which allows model to reflect on the structure between disentangled representations in addition to relationships between samples.
\begin{equation}
\label{combined_loss}
\begin{split}
L_{C}=L_{I} + L_{U} 
\end{split}
\end{equation}

\section{Experiments}
In this section, we focus on the results and analyses of our experiments
for few shot adaptation to shifts in code search. 
To make the best use of the available space, we have included the details 
of the experimental setup—such as the datasets, baselines, and metrics in Appendix F.1$\sim$F.4. 
Furthermore, Appendix F also provides a detailed explanation of the reasons 
behind the occurrence of specific types of shifts (e.g., query shift, code shift, and task shift) 
in each experimental setting.

\begin{table*}[h]
\def\arraystretch{1.0}
\setlength\tabcolsep{8pt} 
\begin{tabular}{@{}lllcccccc@{}}

\toprule
Model                            & \multicolumn{1}{l}{Method}              & \multicolumn{1}{c}{40}           
& \multicolumn{1}{c}{80}         & \multicolumn{1}{c}{120}               & \multicolumn{1}{c}{160}     
& \multicolumn{1}{c}{200}        \\ \midrule

\multirow{2}{*}{CodeT5p-220M}      
& InfoNCE                   & \multicolumn{1}{c}{8.7}             & \multicolumn{1}{c}{13.9}          
                            & \multicolumn{1}{c}{18.8}             & \multicolumn{1}{c}{24.6}          
                            & \multicolumn{1}{c}{24.4}                  
                            \\ \cmidrule(l){2-7} 
& URECA                     & \multicolumn{1}{c}{11.2(+2.6)}          & \multicolumn{1}{c}{16.4(+2.5)}          
                            & \multicolumn{1}{c}{21(+2.2)}          & \multicolumn{1}{c}{26.9(+2.3)}          
                            & \multicolumn{1}{c}{25.6(+1.2)}                   
                            \\ \midrule

\multirow{2}{*}{UniXCoder} 
& InfoNCE                   & \multicolumn{1}{c}{46}             & \multicolumn{1}{c}{48.1}          
                            & \multicolumn{1}{c}{51.8}             & \multicolumn{1}{c}{50.2}          
                            & \multicolumn{1}{c}{50.6}                   
                            \\ \cmidrule(l){2-7} 
& URECA                     & \multicolumn{1}{c}{48.5(+2.4)}          & \multicolumn{1}{c}{51.2(+3.1)}          
                            & \multicolumn{1}{c}{45.1(+2.3)}          & \multicolumn{1}{c}{54.4(+4.2)}          
                            & \multicolumn{1}{c}{54.8(+4.3)}                  
                            \\ \midrule

\multirow{2}{*}{CoCoSoDA} 
& InfoNCE                   & \multicolumn{1}{c}{63.2}             & \multicolumn{1}{c}{63.6}          
                            & \multicolumn{1}{c}{64}             & \multicolumn{1}{c}{63.7}          
                            & \multicolumn{1}{c}{64.7}                    
                            \\ \cmidrule(l){2-7} 
& URECA                     & \multicolumn{1}{c}{64.9(+1.7)}          & \multicolumn{1}{c}{65.8(+2.2)}          
                            & \multicolumn{1}{c}{66.6(+2.6)}          & \multicolumn{1}{c}{67.2(+3.5)}          
                            & \multicolumn{1}{c}{67.7(+3)}                   
                            \\ \bottomrule 
\end{tabular}
\caption{Results of Python across different number of few shot examples (MRR).}
\label{CSN_Python}
\end{table*}

\begin{table*}[h]
\def\arraystretch{1.0}
\setlength\tabcolsep{8pt} 
\begin{tabular}{@{}lllcccccc@{}}

\toprule
Model                            & \multicolumn{1}{l}{Method}              & \multicolumn{1}{c}{40}           
& \multicolumn{1}{c}{80}         & \multicolumn{1}{c}{120}               & \multicolumn{1}{c}{160}     
& \multicolumn{1}{c}{200}        \\ \midrule

\multirow{2}{*}{CodeT5p-220M}      
& InfoNCE                   & \multicolumn{1}{c}{19.7}          & \multicolumn{1}{c}{27}          
                            & \multicolumn{1}{c}{29}          & \multicolumn{1}{c}{33.9}          
                            & \multicolumn{1}{c}{37.7}                  
                            \\ \cmidrule(l){2-7} 
& URECA                     & \multicolumn{1}{c}{23.4(+3.7)}          & \multicolumn{1}{c}{28.2(+1.2)}          
                            & \multicolumn{1}{c}{32.1(+3.1)}          & \multicolumn{1}{c}{36.6(+2.7)}          
                            & \multicolumn{1}{c}{40(+2.3)}                   
                            \\ \midrule

\multirow{2}{*}{UniXCoder} 
& InfoNCE                   & \multicolumn{1}{c}{46.8}          & \multicolumn{1}{c}{45}          
                            & \multicolumn{1}{c}{48}          & \multicolumn{1}{c}{48.5}          
                            & \multicolumn{1}{c}{48.8}                   
                            \\ \cmidrule(l){2-7} 
& URECA                     & \multicolumn{1}{c}{48.2(+1.4)}          & \multicolumn{1}{c}{48.6(+3.6)}          
                            & \multicolumn{1}{c}{48.7(+0.7)}          & \multicolumn{1}{c}{50.3(+1.8)}          
                            & \multicolumn{1}{c}{50.1(+1.3)}                  
                            \\ \midrule

\multirow{2}{*}{CoCoSoDA} 
& InfoNCE                   & \multicolumn{1}{c}{48}          & \multicolumn{1}{c}{49.9}          
                            & \multicolumn{1}{c}{51.6}          & \multicolumn{1}{c}{54.1}          
                            & \multicolumn{1}{c}{54.4}                   
                            \\ \cmidrule(l){2-7} 
& URECA                     & \multicolumn{1}{c}{52(+4)}          & \multicolumn{1}{c}{54.6(+4.7)}          
                            & \multicolumn{1}{c}{54.6(+3)}          & \multicolumn{1}{c}{56.3(+2.2)}          
                            & \multicolumn{1}{c}{55.6(+1.2)}                                  
                            \\ \bottomrule 
\end{tabular}
\caption{Results of CoSQA across different number of few shot examples (MRR).}
\label{CoSQA}
\end{table*}

\subsection{Adaptation to Shifts}
Table 1, 2 and Appendix F.5's Tables show that URECA generally exhibits performance gains over InfoNCE.
This demonstrates the necessity of reflection on disentangled representations to address shifted initialization cascade
across diverse types of shifts.
Experiment section reports results which is average of three different seeds for CSN-Python and CoSQA, 
while results for the remaining programming languages in CSN-Ruby, CSN-Javascript, CSN-Java, CSN-Go and CSN-PHP 
are provided in Appendix F.5.

\textbf{Task Shift: }
UniXCoder has learned patterns of relevance between queries and code through cross-modal generation for generation tasks. 
However, it needs to appropriately adjust these patterns for the code search task, 
leading to the presence of task shift for UniXCoder (detailed information can be found 
in Appendix F.1 to F.4).
In this context, URECA consistently demonstrates performance gains over InfoNCE, 
which means the necessity of leverage for the relationships between disentangled representations.

\textbf{Code Shift: }
Since CodeT5p-220M is a specific version of CodeT5+ 
which has not been pretrained on CSN, both task shift and code shift exist for CSN 
(Details can be found in Appendix F.1 and CSN in Appendix F.2).
Since the current setting is the most harsh, it generally exhibits the lowest performance compared to other settings.
Even under these circumstances, URECA demonstrates consistent performance gains over InfoNCE 
across programming languages and different number of few-shot examples.
These results imply the necessity for leverage of the relationships between disentangled representations

\textbf{Query Shift: }
In the case of CoCoSoDA, it has already been trained on the search task using CodeSearchNet data, 
and the codes in CoSQA also comes from CodeSearchNet. 
Therefore, CoCoSoDA is only related to query shifts for CoSQA.
In constrast to CSN, CoCoSoDA has not been trained on CoSQA (detailed information can be found in Appendix F.2).
Therefore, the performance of CoCoSoDA on CoSQA is meaningful itself unlike CSN in few shot learning  
, and URECA achieves state-of-the-art on CoSQA in the few-shot adaptation setting.
These results are especially important considering that query shifts are the most common type of shifts in real-world scenarios.

Since code of CoSQA is based on CSN-Python code, comparison of the performance improvements in CSN-Python with those in CoSQA allows 
clear understanding for the impact of query shifts. 
In fact, CSN-Python's MRR of URECA for UniXCoder increase by 3.1$\%$ with the increase of few-shot examples,
while  CoSQA's MRR of URECA for UniXCoder increases by only 1.8$\%$ on average. 
This means accurate estimation of dynamics leverages performance of URECA 
since UniXCoder's dynamics of CSN-Python is more accurate than CoSQA due to query shift.

\subsection{Shifted Initialization and Dynamics Estimation}
\begin{table}[h]
\centering
\def\arraystretch{0.8}
\setlength\tabcolsep{6pt} 
\begin{tabular}{@{}lcc@{}}

\toprule
\multicolumn{1}{l}{Method}                      
& \multicolumn{1}{c}{80}          & \multicolumn{1}{c}{120}                \\ \midrule

CSN-Go                      & \multicolumn{1}{c}{43.1 / 46.9 (+3.8)}          
                            & \multicolumn{1}{c}{\textbf{42 / 59.1 (+17.1)}}                          
                            \\ \midrule 
CSN-PHP                     & \multicolumn{1}{c}{\textbf{11.4 / 18 (+6.6)}}          
                            & \multicolumn{1}{c}{15.8 / 17.5 (+1.7)}                             
                            \\ \bottomrule 
\end{tabular}
\caption{Shifted Initialization Cascade of CodeT5+ for CSN-Go/PHP (InfoNCE/URECA(DIFF))}
\label{Shifted Initialization CasCade Performance}
\end{table}
\begin{figure}[h]
\centering
\subfigure[CSN-Go~(Few Shot 120)]{
\includegraphics[width=.45\columnwidth]{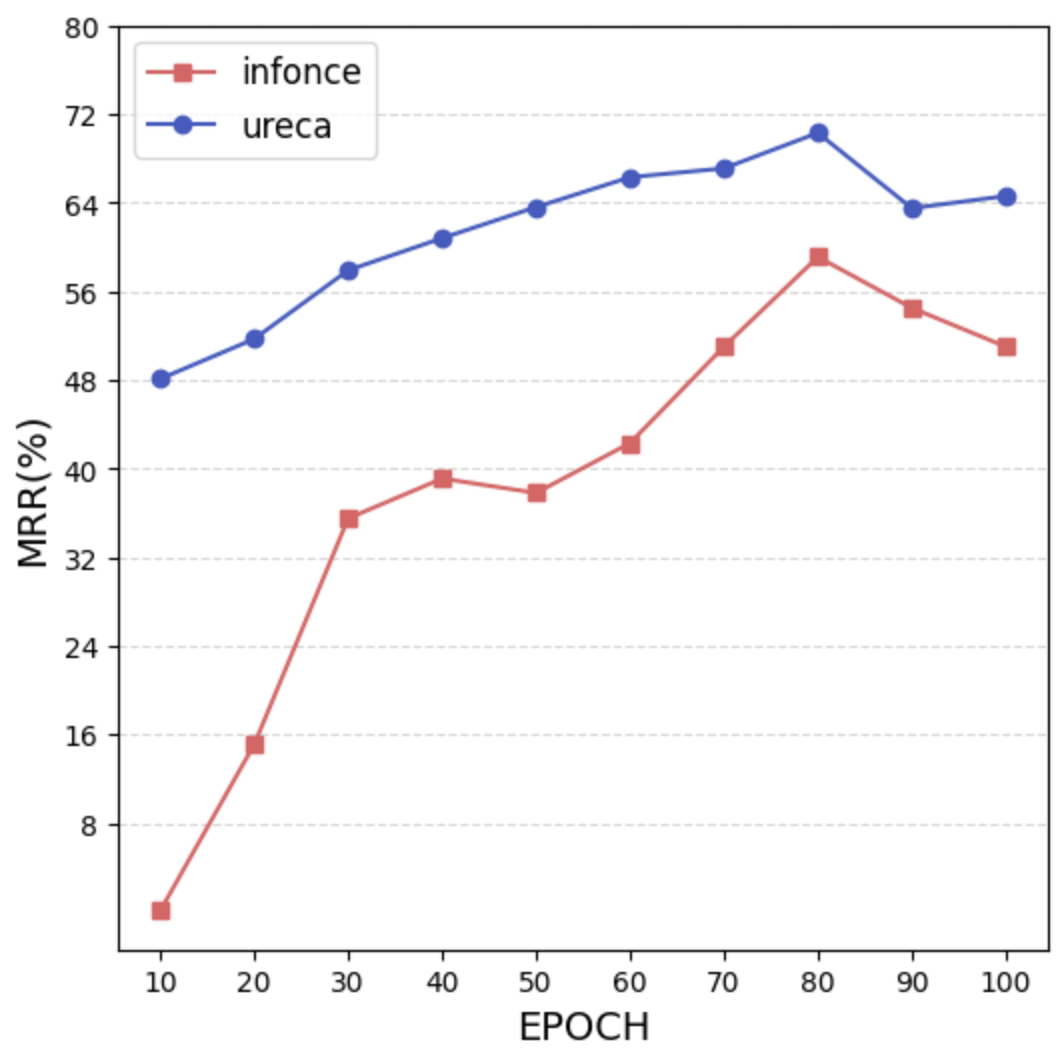}
\label{fig:nrGroup}
}
\subfigure[CSN-PHP~(Few Shot 80)]{
\includegraphics[width=.45\columnwidth]{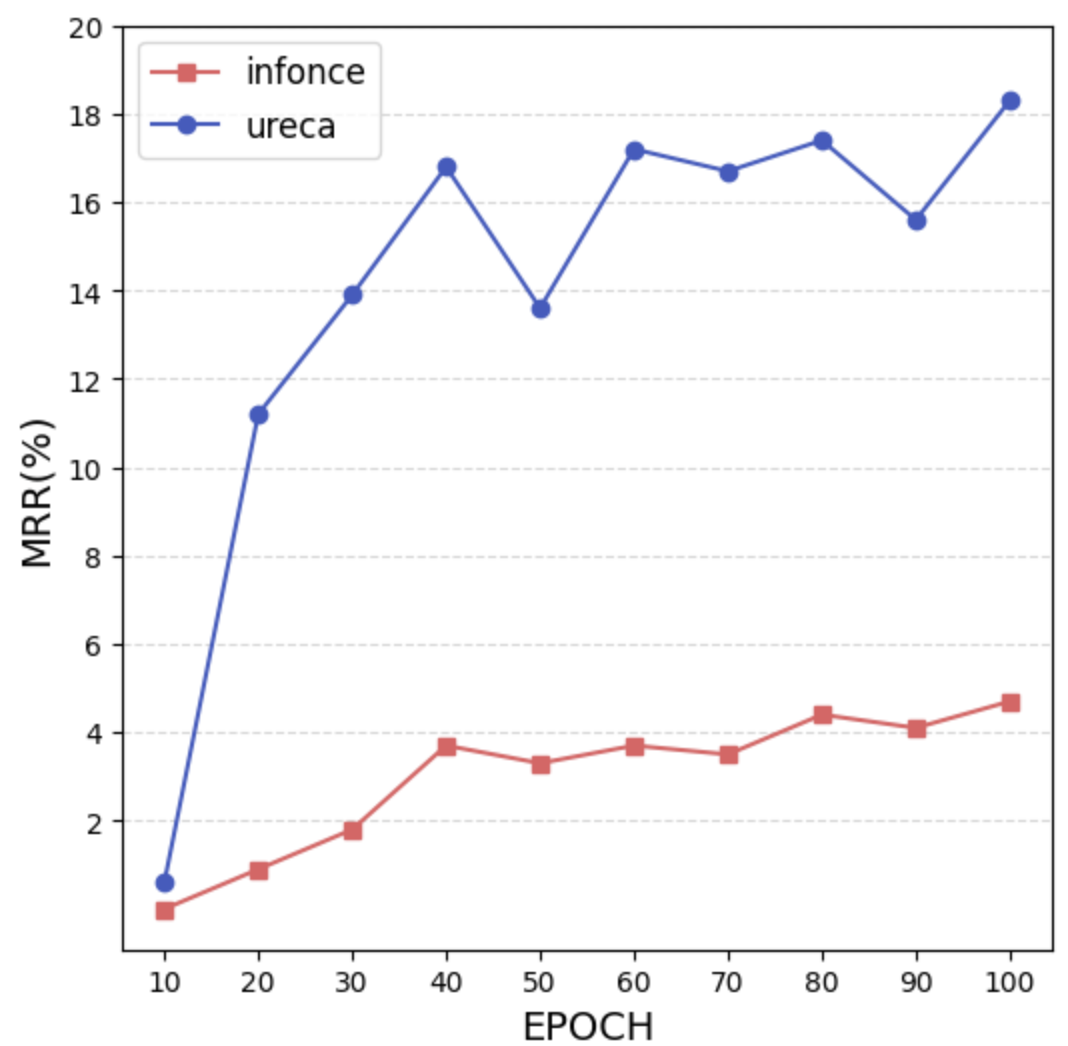}
\label{fig:overallResult}
}
\centering
\caption{
Shifted initialization cascade for CSN-Go and CSN-PHP
}
\end{figure}
\textbf{Shifted Initialization: } 
Figure 3 demonstrates that minimum entropy problem leads to bad solutions due to shifted initialization as mentioned in Section 2.
Figure 3 represents the results of training CodeT5p-220M for 100 epochs with 120 few-shot examples 
of CSN-Go and 80 few-shot examples of CSN-PHP.
For CSN-Go, InfoNCE shows extremely low performance for first 10 epochs (MRR: 0.1$\%$)
while URECA shows superior performance (MRR: 48.1$\%$). 
Although the performance gaps decrease in the process of training,
it is still significant even after 100 epochs for 59.1 $\%$ (InfoNCE) and 70.3$\%$ (URECA). 
For CSN-PHP in Figure 3, InfoNCE still shows little performance improvements again even after 100 epochs of training. 
Although URECA also exhibits little improvements during the first 10 epochs, the performance of URECA spikes after then.
These results demonstrate that 
URECA mitigates the negative effects of shifted initialization with reflection on the relationships between disentangled representations. 
Average performance (11.4 $\%$) in CSN-PHP on Table 3 is higher than performance (4.7 $\%$) for specific seed on Figure 3. 
This shows that the random sampling of few shot examples may release the effect of shifted initialization.
However, since we generally cannot control the suite of few shot examples due to the scarcity of data in real scenario, 
generalization should work robust against the random sampling.
URECA enables this robust generalization in spite of insufficient supervised signals 
based on the reflection of relationships between disentangled representations. 

\textbf{Dynamics Estimation: }
In the case of CoCoSoDA, it has already been trained with supervised signals of relevance based on data augmentation for CSN.
Therefore, CoCoSoDA more accurately estimates the dynamics between disentangled representations for CSN dataset,
which allows us to observe how the precision of estimation for dynamics influences on adaptation to shifts with URECA.
Overall, the performance improvements of URECA in CoCoSoDA are greater than that in UniXCoder, 
which implies that accurate estimation of dynamics is crucial for URECA's performance enhancement.
We also show that Thresholdly Updatable Stationary Assumption outperforms the performance of naive Stationary Assumption 
in Table 11 of Appendix F.6.
These result supports that estimated logits become unbiased estimator for the accurate logits, 
and our estimation based on unit divergence properly works to meets the conditions of Corollary 3.2 for uniform convergence.  

\begin{figure}[t]
\centering
\includegraphics[width=0.98\columnwidth]{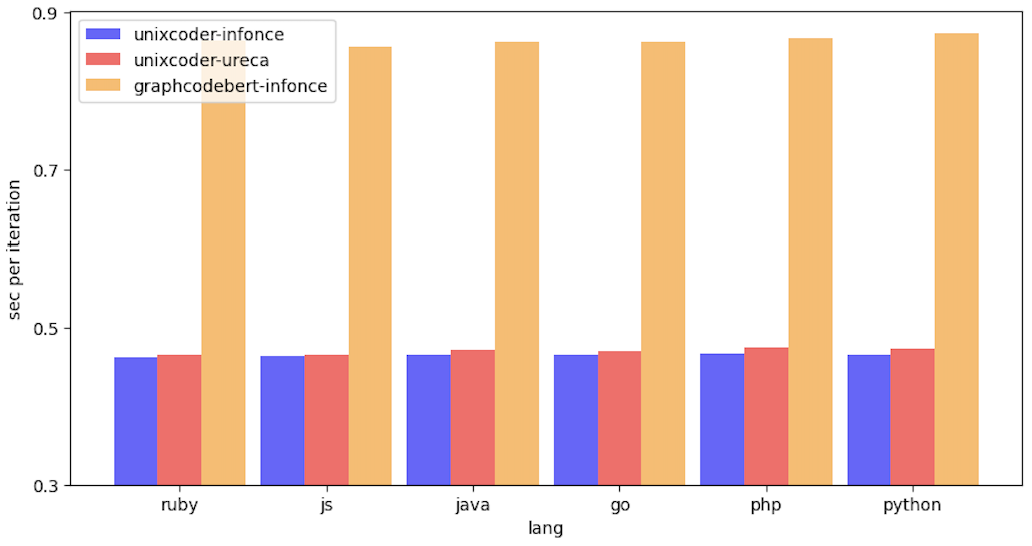}
\caption{Efficiency of Simulation Trick}
\label{efficiency}
\end{figure}

\textbf{Simulation Trick: }We provide the comparison between simulation trick and the approaches of GraphCodeBERT 
(Appendix F.3) in terms of efficiency.
This is especially important since URECA and GraphCodeBERT should be done at every iteration. 
GraphCodeBERT's approaches need heavy computations to guide the relationships between code token 
and node of data flow.
In contrast, simulation trick just needs little computations, 
which leads to large gaps of second per iteration between URECA and GraphCodeBERT in Figure 4.
We also compare the effectiveness and scalability of URECA and GraphCodeBERT as methodologies to leverage semantic structure in Appendix F.7. 
\section{Conclusion}

In this paper, we analyze adaptation as minimum entropy problem. 
We demonstrate that the problem decomposes into the combination of minimum set cover problems, 
which reveals minimum entropy problem ignores the relationships between disentangled representations. 
In this context, we introduce URECA, a clustering algorithm to leverage relationships between disentangled representations. 
Then, we design new auxiliary loss based on the result of URECA to lead the model to reflect on the relationships 
between disentangled representations. 
Our experiments show that URECA achieves SoTA for CoSQA and consistent performance gains across programming languages of CSN. 
These results necessitate to leverage relationships of disentangled representation for robust adaptation, 
paving the way for code retrieval and large-scale RAG systems.

\section{Impact Statement}

This work theoretically analyzes adaptation as minimum entropy problem in terms of Lebesgue integral. 
Minimum entropy problem is one of the most widely used optimization problems across diverse fields 
like statistics, information theory and machine learning). 
This analysis has great potentials to improve data analyses, network communication and machine 
due to the general use of minimum entropy problem 
However, it also brings about the ethical considerations for the privacy about neural network 
since we discover the internal mechanism of minimum entropy problem and connects it to deep learning context.

In addition, our proposed method, URECA leverages the relationships between disentangled representations 
with the clusters estimated from the clustering process of URECA which simulates clustering disentangled 
representations based on simulation trick with transported logits.
This clustering algorithm introduces novel insights for efficient clustering 
since it simulates clustering with numeric calculations.
Although the efficiency makes rapid adaptation to distribution shift, it also accelerates the development
of attack from malicious users.

Although our work pushes the boundaries of machine learning , the distinctions also introduce new challenges
,which people should contemplate on for the proper utilization of advancements.

\addcontentsline{toc}{section}{References}

\bibliographystyle{main}
\bibliography{main}

\begin{thebibliography}{44}
\providecommand{\natexlab}[1]{#1}
\providecommand{\url}[1]{\texttt{#1}}
\expandafter\ifx\csname urlstyle\endcsname\relax
  \providecommand{\doi}[1]{doi: #1}\else
  \providecommand{\doi}{doi: \begingroup \urlstyle{rm}\Url}\fi

\bibitem[Arakelyan et~al.(2023)Arakelyan, Das, Mao, and Ren]{ArakelyanDMR23}
Arakelyan, S., Das, R.~J., Mao, Y., and Ren, X.
\newblock Exploring distributional shifts in large language models for code analysis.
\newblock In \emph{Proceedings of the 2023 Conference on Empirical Methods in Natural Language Processing, {EMNLP} 2023, Singapore, December 6-10, 2023}, pp.\  16298--16314, 2023.

\bibitem[Asadi et~al.(2018)Asadi, Misra, and Littman]{KavoshDM18}
Asadi, K., Misra, D., and Littman, M.
\newblock {L}ipschitz continuity in model-based reinforcement learning.
\newblock In Dy, J. and Krause, A. (eds.), \emph{Proceedings of the 35th International Conference on Machine Learning}, volume~80 of \emph{Proceedings of Machine Learning Research}, pp.\  264--273. PMLR, 10--15 Jul 2018.
\newblock URL \url{https://proceedings.mlr.press/v80/asadi18a.html}.

\bibitem[Brown et~al.(2020)Brown, Mann, Ryder, Subbiah, Kaplan, Dhariwal, Neelakantan, Shyam, Sastry, Askell, et~al.]{BrownBNMJPA20}
Brown, T., Mann, B., Ryder, N., Subbiah, M., Kaplan, J.~D., Dhariwal, P., Neelakantan, A., Shyam, P., Sastry, G., Askell, A., et~al.
\newblock Language models are few-shot learners.
\newblock \emph{Advances in neural information processing systems}, 33:\penalty0 1877--1901, 2020.

\bibitem[Cambronero et~al.(2019)Cambronero, Li, Kim, Sen, and Chandra]{CambroneroLKS019}
Cambronero, J., Li, H., Kim, S., Sen, K., and Chandra, S.
\newblock When deep learning met code search.
\newblock In \emph{Proceedings of the {ACM} Joint Meeting on European Software Engineering Conference and Symposium on the Foundations of Software Engineering, {ESEC/SIGSOFT} {FSE}}, pp.\  964--974, 2019.

\bibitem[Cardinal et~al.(2012)Cardinal, Fiorini, and Joret]{CardinalFJ12}
Cardinal, J., Fiorini, S., and Joret, G.
\newblock Minimum entropy combinatorial optimization problems.
\newblock \emph{Theory of Computing Systems}, 51:\penalty0 4--21, 2012.

\bibitem[Chen et~al.(2024)Chen, Xiong, Liu, Wu, Xiao, Gao, and He]{ShiqiMJZTSJ24}
Chen, S., Xiong, M., Liu, J., Wu, Z., Xiao, T., Gao, S., and He, J.
\newblock In-context sharpness as alerts: An inner representation perspective for hallucination mitigation.
\newblock In Salakhutdinov, R., Kolter, Z., Heller, K., Weller, A., Oliver, N., Scarlett, J., and Berkenkamp, F. (eds.), \emph{Proceedings of the 41st International Conference on Machine Learning}, volume 235 of \emph{Proceedings of Machine Learning Research}, pp.\  7553--7567. PMLR, 21--27 Jul 2024.
\newblock URL \url{https://proceedings.mlr.press/v235/chen24av.html}.

\bibitem[Chen et~al.(2016)Chen, Duan, Houthooft, Schulman, Sutskever, and Abbeel]{ChenDHSSA16}
Chen, X., Duan, Y., Houthooft, R., Schulman, J., Sutskever, I., and Abbeel, P.
\newblock Infogan: Interpretable representation learning by information maximizing generative adversarial nets.
\newblock \emph{Advances in neural information processing systems}, 29, 2016.

\bibitem[Denton(2017)]{Denton17}
Denton, E.~L.
\newblock Unsupervised learning of disentangled representations from video.
\newblock \emph{Advances in neural information processing systems}, 30, 2017.

\bibitem[Feng et~al.(2020)Feng, Guo, Tang, Duan, Feng, Gong, Shou, Qin, Liu, Jiang, and Zhou]{FengGTDFGS0LJZ20}
Feng, Z., Guo, D., Tang, D., Duan, N., Feng, X., Gong, M., Shou, L., Qin, B., Liu, T., Jiang, D., and Zhou, M.
\newblock Codebert: {A} pre-trained model for programming and natural languages.
\newblock In \emph{Findings of the Association for Computational Linguistics: {EMNLP}}, pp.\  1536--1547, 2020.

\bibitem[Grandvalet \& Bengio(2004)Grandvalet and Bengio]{Yvew04}
Grandvalet, Y. and Bengio, Y.
\newblock Semi-supervised learning by entropy minimization.
\newblock \emph{Advances in neural information processing systems}, 17, 2004.

\bibitem[Greiff(2000)]{Greiff00}
Greiff, W.~R.
\newblock \emph{The use of Exploratory Data Analysis in Information Retrieval Research}, pp.\  37--72.
\newblock Springer US, Boston, MA, 2000.
\newblock \doi{10.1007/0-306-47019-5_2}.
\newblock URL \url{https://doi.org/10.1007/0-306-47019-5_2}.

\bibitem[Gu et~al.(2018)Gu, Zhang, and Kim]{GuZ018}
Gu, X., Zhang, H., and Kim, S.
\newblock Deep code search.
\newblock In \emph{Proceedings of the 40th International Conference on Software Engineering, {ICSE}}, pp.\  933--944, 2018.

\bibitem[Guo et~al.(2021)Guo, Ren, Lu, Feng, Tang, Liu, Zhou, Duan, Svyatkovskiy, Fu, Tufano, Deng, Clement, Drain, Sundaresan, Yin, Jiang, and Zhou]{GuoRLFT0ZDSFTDC21}
Guo, D., Ren, S., Lu, S., Feng, Z., Tang, D., Liu, S., Zhou, L., Duan, N., Svyatkovskiy, A., Fu, S., Tufano, M., Deng, S.~K., Clement, C.~B., Drain, D., Sundaresan, N., Yin, J., Jiang, D., and Zhou, M.
\newblock {GraphCodeBERT}: Pre-training code representations with data flow.
\newblock In \emph{9th International Conference on Learning Representations, {ICLR}}, 2021.

\bibitem[Guo et~al.(2022)Guo, Lu, Duan, Wang, Zhou, and Yin]{GuoLDW0022}
Guo, D., Lu, S., Duan, N., Wang, Y., Zhou, M., and Yin, J.
\newblock {UniXcoder}: Unified cross-modal pre-training for code representation.
\newblock In \emph{Proceedings of the 60th Annual Meeting of the Association for Computational Linguistics, {ACL}}, pp.\  7212--7225, 2022.

\bibitem[Halim \& Gilbert(2016)Halim and Gilbert]{HalimG16}
Halim, S. and Gilbert, S.
\newblock Min-set-cover, August 2016.

\bibitem[Halperin \& Karp(2005)Halperin and Karp]{HalperinK05}
Halperin, E. and Karp, R.~M.
\newblock The minimum-entropy set cover problem.
\newblock \emph{Theoretical Computer Science}, 348:\penalty0 240--250, 2005.

\bibitem[Huang et~al.(2021)Huang, Tang, Shou, Gong, Xu, Jiang, Zhou, and Duan]{HuangTSG0J0D20}
Huang, J., Tang, D., Shou, L., Gong, M., Xu, K., Jiang, D., Zhou, M., and Duan, N.
\newblock {CoSQA}: 20, 000+ web queries for code search and question answering.
\newblock In \emph{Proceedings of the 59th Annual Meeting of the Association for Computational Linguistics and the 11th International Joint Conference on Natural Language Processing, {ACL/IJCNLP}}, pp.\  5690--5700, 2021.

\bibitem[Huang et~al.(2025)Huang, Yu, Ma, Zhong, Feng, Wang, Chen, Peng, Feng, Qin, and Liu]{LeiWWZHQWXBT25}
Huang, L., Yu, W., Ma, W., Zhong, W., Feng, Z., Wang, H., Chen, Q., Peng, W., Feng, X., Qin, B., and Liu, T.
\newblock A survey on hallucination in large language models: Principles, taxonomy, challenges, and open questions.
\newblock \emph{ACM Trans. Inf. Syst.}, 43\penalty0 (2), January 2025.
\newblock ISSN 1046-8188.
\newblock \doi{10.1145/3703155}.
\newblock URL \url{https://doi.org/10.1145/3703155}.

\bibitem[Husain et~al.(2019)Husain, Wu, Gazit, Allamanis, and Brockschmidt]{HusainWGAB19}
Husain, H., Wu, H., Gazit, T., Allamanis, M., and Brockschmidt, M.
\newblock {CodeSearchNet Challenge}: Evaluating the state of semantic code search.
\newblock \emph{ArXiV CoRR}, 1909.09436, 2019.

\bibitem[Jeon et~al.(2021)Jeon, Lee, Pyeon, and Kim]{JeonLPK21}
Jeon, I., Lee, W., Pyeon, M., and Kim, G.
\newblock Ib-gan: Disentangled representation learning with information bottleneck generative adversarial networks.
\newblock In \emph{Proceedings of the AAAI conference on artificial intelligence}, volume~35, pp.\  7926--7934, 2021.

\bibitem[Kwon \& Yoon(2012)Kwon and Yoon]{KwonY12}
Kwon, S. and Yoon, H.
\newblock Lebesgue integral theory, 2012.

\bibitem[Lee et~al.(2024)Lee, Jung, Lee, Park, Shin, Hwang, and Yoon]{LeeJLPSHY24}
Lee, J., Jung, D., Lee, S., Park, J., Shin, J., Hwang, U., and Yoon, S.
\newblock Entropy is not enough for test-time adaptation: From the perspective of disentangled factors.
\newblock In \emph{The Twelfth International Conference on Learning Representations, {ICLR}}, 2024.

\bibitem[Lewis et~al.(2020)Lewis, Perez, Piktus, Petroni, Karpukhin, Goyal, K{\"u}ttler, Lewis, Yih, Rockt{\"a}schel, et~al.]{LewisEAFVNHMWT20}
Lewis, P., Perez, E., Piktus, A., Petroni, F., Karpukhin, V., Goyal, N., K{\"u}ttler, H., Lewis, M., Yih, W.-t., Rockt{\"a}schel, T., et~al.
\newblock Retrieval-augmented generation for knowledge-intensive nlp tasks.
\newblock \emph{Advances in Neural Information Processing Systems}, 33:\penalty0 9459--9474, 2020.

\bibitem[Li et~al.(2023)Li, Zhou, Luu, and Miao]{HaochenXAC23}
Li, H., Zhou, X., Luu, A., and Miao, C.
\newblock Rethinking negative pairs in code search.
\newblock In Bouamor, H., Pino, J., and Bali, K. (eds.), \emph{Proceedings of the 2023 Conference on Empirical Methods in Natural Language Processing}, pp.\  12760--12774, Singapore, December 2023. Association for Computational Linguistics.
\newblock \doi{10.18653/v1/2023.emnlp-main.786}.
\newblock URL \url{https://aclanthology.org/2023.emnlp-main.786/}.

\bibitem[Li et~al.(2024)Li, Zhou, and Shen]{Haochen24}
Li, H., Zhou, X., and Shen, Z.
\newblock Rewriting the code: A simple method for large language model augmented code search.
\newblock In Ku, L.-W., Martins, A., and Srikumar, V. (eds.), \emph{Proceedings of the 62nd Annual Meeting of the Association for Computational Linguistics (Volume 1: Long Papers)}, pp.\  1371--1389, Bangkok, Thailand, August 2024. Association for Computational Linguistics.
\newblock \doi{10.18653/v1/2024.acl-long.75}.
\newblock URL \url{https://aclanthology.org/2024.acl-long.75/}.

\bibitem[Lin et~al.(2019)Lin, Thekumparampil, Fanti, and Oh]{LinKFO19}
Lin, Z., Thekumparampil, K.~K., Fanti, G., and Oh, S.
\newblock Infogan-cr: Disentangling generative adversarial networks with contrastive regularizers.
\newblock \emph{arXiv preprint arXiv:1906.06034}, pp.\ ~60, 2019.

\bibitem[Liu et~al.(2021)Liu, Li, Niu, Xu, and Zhang]{LiuLNXZ21}
Liu, L., Li, J., Niu, L., Xu, R., and Zhang, L.
\newblock Activity image-to-video retrieval by disentangling appearance and motion.
\newblock In \emph{Proceedings of the AAAI Conference on Artificial Intelligence}, volume~35, pp.\  2145--2153, 2021.

\bibitem[Madhulatha(2012)]{Soni12}
Madhulatha, T.~S.
\newblock An overview on clustering methods.
\newblock \emph{arXiv preprint arXiv:1205.1117}, 2012.

\bibitem[Mao et~al.(2023)Mao, Mohri, and Zhong]{AnqiMY23}
Mao, A., Mohri, M., and Zhong, Y.
\newblock Cross-entropy loss functions: Theoretical analysis and applications.
\newblock In \emph{International conference on Machine learning}, pp.\  23803--23828. PMLR, 2023.

\bibitem[Niu et~al.(2024)Niu, Wu, Zhu, Xu, Shum, Zhong, Song, and Zhang]{ChenYJSKRJT24}
Niu, C., Wu, Y., Zhu, J., Xu, S., Shum, K., Zhong, R., Song, J., and Zhang, T.
\newblock {RAGT}ruth: A hallucination corpus for developing trustworthy retrieval-augmented language models.
\newblock In Ku, L.-W., Martins, A., and Srikumar, V. (eds.), \emph{Proceedings of the 62nd Annual Meeting of the Association for Computational Linguistics (Volume 1: Long Papers)}, pp.\  10862--10878, Bangkok, Thailand, August 2024. Association for Computational Linguistics.
\newblock \doi{10.18653/v1/2024.acl-long.585}.
\newblock URL \url{https://aclanthology.org/2024.acl-long.585/}.

\bibitem[Oord et~al.(2018)Oord, Li, and Vinyals]{DenYO18}
Oord, A. v.~d., Li, Y., and Vinyals, O.
\newblock Representation learning with contrastive predictive coding.
\newblock \emph{arXiv preprint arXiv:1807.03748}, 2018.

\bibitem[Press et~al.(2024)Press, Shwartz-Ziv, LeCun, and Bethge]{OriRYM24}
Press, O., Shwartz-Ziv, R., LeCun, Y., and Bethge, M.
\newblock The entropy enigma: Success and failure of entropy minimization.
\newblock In Salakhutdinov, R., Kolter, Z., Heller, K., Weller, A., Oliver, N., Scarlett, J., and Berkenkamp, F. (eds.), \emph{Proceedings of the 41st International Conference on Machine Learning}, volume 235 of \emph{Proceedings of Machine Learning Research}, pp.\  41064--41085. PMLR, 21--27 Jul 2024.

\bibitem[Shannon(1948)]{Shannon48}
Shannon, C.~E.
\newblock A mathematical theory of communication.
\newblock \emph{Bell System Technical Journal}, 27:\penalty0 379--423, 1948.

\bibitem[Shi et~al.(2023)Shi, Wang, Gu, Du, Zhang, Han, Zhang, and Sun]{EnshenYWLHSDH23}
Shi, E., Wang, Y., Gu, W., Du, L., Zhang, H., Han, S., Zhang, D., and Sun, H.
\newblock Cocosoda: Effective contrastive learning for code search.
\newblock In \emph{Proceedings of the 45th International Conference on Software Engineering}, ICSE '23, pp.\  2198–2210. IEEE Press, 2023.
\newblock ISBN 9781665457019.
\newblock \doi{10.1109/ICSE48619.2023.00185}.
\newblock URL \url{https://doi.org/10.1109/ICSE48619.2023.00185}.

\bibitem[Tran et~al.(2017)Tran, Yin, and Liu]{TranYL17}
Tran, L., Yin, X., and Liu, X.
\newblock Disentangled representation learning gan for pose-invariant face recognition.
\newblock In \emph{Proceedings of the IEEE conference on computer vision and pattern recognition}, pp.\  1415--1424, 2017.

\bibitem[Wang et~al.(2021{\natexlab{a}})Wang, Shelhamer, Liu, Olshausen, and Darrell]{WangSLOD21}
Wang, D., Shelhamer, E., Liu, S., Olshausen, B., and Darrell, T.
\newblock Tent: Fully test-time adaptation by entropy minimization.
\newblock In \emph{International Conference on Learning Representations}, 2021{\natexlab{a}}.
\newblock URL \url{https://openreview.net/forum?id=uXl3bZLkr3c}.

\bibitem[Wang et~al.(2022)Wang, Chen, Tang, Wu, and Zhu]{WangCTWZ22}
Wang, X., Chen, H., Tang, S., Wu, Z., and Zhu, W.
\newblock Disentangled representation learning.
\newblock \emph{arXiv preprint arXiv:2211.11695}, 2022.

\bibitem[Wang et~al.(2021{\natexlab{b}})Wang, Wang, Joty, and Hoi]{WangWJH21}
Wang, Y., Wang, W., Joty, S.~R., and Hoi, S. C.~H.
\newblock {CodeT5}: Identifier-aware unified pre-trained encoder-decoder models for code understanding and generation.
\newblock In \emph{Proceedings of the 2021 Conference on Empirical Methods in Natural Language Processing, {EMNLP}}, pp.\  8696--8708, 2021{\natexlab{b}}.

\bibitem[Wang et~al.(2023)Wang, Le, Gotmare, Bui, Li, and Hoi]{WangLGB0H23}
Wang, Y., Le, H., Gotmare, A., Bui, N. D.~Q., Li, J., and Hoi, S. C.~H.
\newblock {CodeT5+}: Open code large language models for code understanding and generation.
\newblock In \emph{Proceedings of the 2023 Conference on Empirical Methods in Natural Language Processing, {EMNLP}}, pp.\  1069--1088, 2023.

\bibitem[Wiles et~al.(2022)Wiles, Gowal, Stimberg, Rebuffi, Ktena, Dvijotham, and Cemgil]{WilesGSRKDC22}
Wiles, O., Gowal, S., Stimberg, F., Rebuffi, S., Ktena, I., Dvijotham, K., and Cemgil, A.~T.
\newblock A fine-grained analysis on distribution shift.
\newblock In \emph{The Tenth International Conference on Learning Representations, {ICLR}}, 2022.

\bibitem[Xiao et~al.(2017)Xiao, Hong, and Ma]{XiaoHM17}
Xiao, T., Hong, J., and Ma, J.
\newblock Dna-gan: Learning disentangled representations from multi-attribute images.
\newblock \emph{arXiv preprint arXiv:1711.05415}, 2017.

\bibitem[Yin et~al.(2018)Yin, Deng, Chen, Vasilescu, and Neubig]{PengchenBEBG18}
Yin, P., Deng, B., Chen, E., Vasilescu, B., and Neubig, G.
\newblock Learning to mine aligned code and natural language pairs from stack overflow.
\newblock In \emph{Proceedings of the 15th international conference on mining software repositories}, pp.\  476--486, 2018.

\bibitem[Zhang et~al.(2021)Zhang, Hong, Zhang, Wan, Liu, and Sui]{ZhangHZWLS21}
Zhang, J., Hong, H., Zhang, Y., Wan, Y., Liu, Y., and Sui, Y.
\newblock Disentangled code representation learning for multiple programming languages.
\newblock \emph{Findings of the Association for Computational Linguistics: ACL-IJCNLP 2021}, 2021.

\bibitem[Zhou et~al.(2023)Zhou, Alon, Xu, Jiang, and Neubig]{Zhou0XJN23}
Zhou, S., Alon, U., Xu, F.~F., Jiang, Z., and Neubig, G.
\newblock Docprompting: Generating code by retrieving the docs.
\newblock In \emph{The Eleventh International Conference on Learning Representations, {ICLR}}, 2023.

\end{thebibliography}

\onecolumn
\appendix
\section{Related Works}

\subsection{Disentangled Representation Learning}
Disentangled representation learning has emerged as a critical area of study within the broader field of representation learning.
This can be categorized into dimension-wise and vector-wise approaches according to representation structure~(\cite{WangCTWZ22}).
Dimension-wise methods assign a single scalar dimension to represent fine-grained generative factors, 
making them suitable for synthetic and simple datasets~(\cite{ChenDHSSA16, JeonLPK21, LinKFO19, XiaoHM17}).
In contrast, vector-wise methods use multiple dimensions (vectors) to represent coarse-grained factors, 
making them more applicable to complex real-world tasks such as identity swapping, 
image classification and video understanding~(\cite{TranYL17, LiuLNXZ21, Denton17, WangCTWZ22}).
In this paper, we focus on vector-wise approaches 
since flexible approaches are suitable to code search as one of the most realistic tasks.

(\cite{WilesGSRKDC22, LeeJLPSHY24}) explore the disentangled representations and (\cite{LeeJLPSHY24}) highlights 
that minimum entropy problem ignores the influence of disentangled representations, 
which leads to unreliable prediction.
In addition, (\cite{WilesGSRKDC22}) assists that it is important 
for model to understand disentangled representations for distribution shift.
However, most methodologies for disentangled representation learning do not focus on 
the learning process itself which drives model to learn entanglement of fine-grained representations
In contrast, this paper analyzes adaptation as minimum entropy problem in terms of Lebesgue integral 
and proposes a new algorithm for clustering disentangled representations 
to reflect on the internal structure of representation

\subsection{Semantic Code Search}
Semantic Code Search~(\cite{GuZ018,CambroneroLKS019}) is a task to retrieve the most relevant code snippets 
in response to given natural language query.
This semantic code search is especially useful for hallucination of LLM with leverage for the result of 
retrieval~(\cite{ChenYJSKRJT24,ShiqiMJZTSJ24}). 
It has significantly evolved in accordance with the advancements in deep learning.
Some datasets such as CodeSearchNet~(\cite{HusainWGAB19}) and CoSQA~(\cite{HuangTSG0J0D20}) are introduced 
and other foundational works did significant efforts to advance the frontiers to the next level
~(\cite{EnshenYWLHSDH23, HaochenXAC23, Haochen24}).
These datasets not only support the development of code-specific pretrained language model 
~(\cite{GuoRLFT0ZDSFTDC21, FengGTDFGS0LJZ20, WangWJH21, GuoLDW0022, WangLGB0H23}) but also enable the fine-tuning of these models, 
leading to notable improvements in code search tasks~(\cite{GuoLDW0022, WangLGB0H23}). 
GraphCodeBERT~(\cite{FengGTDFGS0LJZ20}), UniXCoder~(\cite{GuoLDW0022}) and CodeT5+~(\cite{WangLGB0H23}) stand out 
as these programming language models that are the ones of the most successful models in code search task.

As long as we know, there is no single paper that 
treats the distribution shift in code search except for docprompting~(\cite{Zhou0XJN23}) 
which only partially treats code search under the setting of distribution shift 
since the authors focus on the code generation.
However, the retrieval module is enough deserved to be spotlighted, 
considering the sensitivity of LLMs to given information and the hallucination as the problem deep-seated into LLMs.
Disentangled representation learning is well-suited to the solution of this distribution shift, 
especially in the domain of programming language 
in that PL is structured language and pattern in this domain is sensitive to the structure of disentangled fragments.
This learning paradigm is out of sight of the researchers in code domain 
except for (\cite{ZhangHZWLS21}) which try to learn disentangled representations in code domain for multilingual generalization.
However, they do only focus on code translation not on code search 
in spite of the intimate connection between code search and the representation learning.
In contrast to these previous efforts, 
we focus on disentangled representation learning as an adaptation paradigm 
to diverse types of shifts in code search for the leverage of structure in disentangled representations.

\subsection{Minimum Entropy Problem}
Entropy is originally introduced as optimal length of code by (\cite{Shannon48, AnqiMY23}).
This entropy is formally defined as the expectation of self-information across different events.
The self-information can be interpreted as a measure for the amount of information 
since it is inversely proportional to the possibility of specific event.
If that event is likely to happen, the occurrence of that event does not update much references for decision.
In contrast, the occurrence of unlikely event updates many parts of codebooks.
These correlations makes self-information proportional to the amount of updates, 
which is the reason why we use this measure for information.

Minimum entropy problem is the problem to minimize the expectation 
of this amount of updates by exploring the space of parameters.
This problem is widely used in the field of machine learning for classification with cross entropy.
In addition, entropy minimization plays a crucial role in Test Time Adaptation 
that can improve the generalization to shifted distributions in unsupervised manner(\cite{WangSLOD21}). 
(\cite{OriRYM24}) provides experimental supports for the hypothesis that entropy minimization is related to clustering.
Unlike (\cite{WangSLOD21}) and (\cite{OriRYM24}), we theoretically shed on the light 
into the internal mechanism of minimum entropy problem 
which can be used to train model regardless of the existence of supervised signals.
This makes us realize the reason that shifted initialization results in the bad solution of clustering.

The minimum set cover problem is one of the most fundamental optimization problems, which is NP-complete.
Given universe $U$, a collection $T$ as a power set of $U$, and cost function $c$, 
this problem is defined to find a sub-collection of $T$ whose union of all elements covers $U$ whose cost is minimum.
Greedy algorithm selects subset $s$ whose price is minimum for each iteration until all elements of $U$ are covered.
It is well known that this algorithm becomes best approximation unless NP has polynomial time algorithm.
This means that the minimum set cover as solution constructed by greedy algorithm has tight upper bound 
for the cost, $c(OPT)(\ln{1 \over p}+\gamma)$ when $\gamma$ is Euler-Mascheroni constant 
and $OPT$ is the optimal solution of the minimum set cover problem.

According to (\cite{HalperinK05, CardinalFJ12}), the minimum set cover problem is dual form of the minimum entropy problem, 
which is called the minimum entropy set cover problem.
However, existing researches only focus on the superficial parts of this problem.
With the analysis of the minimum set cover problem in terms of Lebesgue integral, 
we extend the connection of the minimum set cover problem and minimum entropy problem into the unknowns.
We unveil another minimum set cover problem behind the minimum entropy set cover problem.
\section{Lebesgue Integral}
Lebesgue integral is extension of riemannian integral(\cite{KwonY12}). 
Intuitively, lebesgue integral is integral across the partitions of image 
for the given function, while riemannian integral is across the partitions of domain. 
Lebesgue integral for the non-negative measurable function $f$ is defined as follows,
under the condition $0 \le s(x) \le f(x)$.

\begin{equation}
\label{APP_B_def:lebesgue_integral}
\int{f(x)  d\mu(x)} = \sup\{\int  s(x) d\mu(x)\}
\end{equation}
In this formula, $s$ is measurable simple function, which is linear combination of characteristic function $\chi_{E_k}$. 
$a_k$ is the element in range  and $E_k$ is pre-image of $s$.
Measurable function is function of which all pre-images are in the domain of that function.  
Definition of simple function is as follows.
\begin{equation}
\label{def:simple_function}
s(x) = \sum_k a_k \chi_{E_k}(x) 
\end{equation}
Characteristic function $\chi_{E_k}$ can be called as indicator function, which is defined as follows. 
\begin{align}
\label{def:charac_func}
\chi_{E_k}(x) = 
     \begin{cases}
       1 & (x \in E_k)\\
       0 & (x \notin E_k)\\
     \end{cases}
\end{align}
\section{Pseudo Code of URECA}
\begin{algorithm}
\caption{URECA}
\label{alg:algorithm}
\textbf{Input}:  Q, D, GT \\
\textbf{Output}: NONE
\begin{algorithmic}[1]
\State $E=Q \cdot D$
\State $DYN=softmax(Q \cdot Q)$
\For{$k \gets 1$ to $max\_recursion\_num$}
\State $S = argsort(E)$
\State $p = S.find\_position(GT)$
\State $F, B = E.split(p)$
\State $F = F + F \cdot DYM $
\State $E=F$
\EndFor
\end{algorithmic}
\end{algorithm}





\section{Proofs}
\subsection{Proof for Theorem 2.1}
The outline of proof is as follows. 
At first, we prove the base case of Theorem 2.1 from the relationship of self-information and the minimum set cover problem which is described by Lebesgue integral.
Base case means the probability of each element in universe is ${1 \over 2^k}$ and we extends the relationship 
to general case in terms of probability and the cardinality of universe.
At last, we illuminate the relationship between minimum entropy problem and the minimum set cover problem from self-information’s one.

\begin{proof} 
For the first, we explain the basics of the minimum set cover problem (\cite{HalimG16}).
The minimum set cover problem is one of the most fundamental optimization problems, which is NP-complete. 
Given universe $U_\alpha$ , a collection $T$ as a power set of $U_\alpha$($\alpha$ is the identifier for specific event) 
, and cost function $c$, this problem is defined to find a sub-collection of $T$ whose union of all elements covers $U_{\alpha}$ 
and the cost is minimum. 
Greedy algorithm selects subsets whose price is minimum for each iteration until all elements of $U_{\alpha}$ are covered.
According to (\cite{HalimG16}), this greedy algorithm constructs the minimum set cover and 
the tight upper bound of cost function for this solution is $\ln n$ 
when $n$ is the cardinality of universe $U_{\alpha}$.
Assume that $OPT$ is the fixed optimal set cover and $u^{\alpha}_1,u^{\alpha}_2,...,u^{\alpha}_{n}$ 
(elements of $U_{\alpha}$) are sorted in the covered order by greedy algorithm.
If we consider that $C_{k-1}=\{u^{\alpha}_1,u^{\alpha}_2,...,u^{\alpha}_{k-1}\}$ has been covered already, 
then the collection of optimal sets selected by greedy algorithm from now on is $OPT_k$=$\{O_1, O_2,...,O_r\}$. 
For each $O_i$ as a subset of $U_{\alpha}$, the lower bound of residual elements is described by equation (21),
which the equality holds true when all $O_i$s are disjoint to each other. 
\begin{equation}
\label{lower_bound_msc}
\sum_{i=1}^r |O_i\cap(U_{\alpha}\backslash C_{k-1})| 
\ge |U_{\alpha}\backslash C_{k-1}|
\end{equation}

\begin{equation}
\label{cardinality_residual_candi}
|U_{\alpha}\backslash C_{k-1}| = n-k+1
\end{equation}

By definition, the $price\_per\_item$ (price for each element of universe) is as follows, 
for each $j \in \{1,2,...,r\}$.
\begin{equation}
\label{price_per_item_selection}
price\_per\_item(O_j)=\frac{c(O_j)}{|O_j \cap (U_{\alpha} \backslash C_{k-1})|}
\end{equation}

Since the optimal greedy algorithm selects the optimal set whose $price\_per\_item$ is minimum,
the upper bound of price of $e_k$ is as follows.  
\begin{align}
\label{up_bound_price_per_item}
price(u^{\alpha}_k) \le \frac{c(O_j)}{|O_j \cap (U_{\alpha} \backslash C_{k-1})|} 
\end{align}
The entire cost of $OPT_k$ is as follows.
\begin{equation}
\label{low_bound_cost_optimum}
\begin{split}
c(OPT_k) &= \sum_{j=1}^rc(O_j)\\
         &\ge price(u^{\alpha}_k) \cdot \sum_{j=1}^r|O_j \cap (U_{\alpha}\backslash C_{k-1})|\\
         &\ge price(u^{\alpha}_k) \cdot |U_{\alpha} \backslash C_{k-1}|\\
         &= price(u^{\alpha}_k) \cdot (n-k+1)
\end{split}
\end{equation}

Let $V$ be a collection of subsets selected by greedy algorithm. 
Then the upper bound for cost of $V$ is as follows.
\begin{equation}
\label{up_bound_cost}
\begin{split}
c(V) &= \sum_{k=1}^n price(u^{\alpha}_k) \\
     &\le \sum_{k=1}^n\frac{c(OPT_k)}{n-k+1} \\
     &\le \sum_{k=1}^n\frac{c(OPT)}{n-k+1} \\
     &\le c(OPT) \cdot \sum_{k=1}^n \frac{1}{n-k+1} \\
     &= c(OPT) \cdot \sum_{k=1}^n \frac{1}{k} \\
     &\approx c(OPT)(\ln n + \gamma)
\end{split}
\end{equation} 

Now we will describe self-information in terms of Lebesgue integral like equation (1) in Section 2 
and connect the description with the minimum set cover problem.
\begin{align}
\label{lebesgue_integral_entropy}
\begin{split}
\ln {1 \over p(E_\alpha)} &=\int p(E_\alpha)d{1 \over p(E_\alpha)}\\
                          &=\sup\{\sum_{n=1}^{\lceil{1 \over p(E_{\alpha})}\rceil}p(E_{\alpha}) \cdot 1\}\\  
\end{split}
\end{align}
Based on equation (26) and details about minimum set cover, 
we interpret $\sum_{n=1}^{\lceil{1 \over p(E_\alpha)}\rceil}p(E_\alpha)\cdot 1$ as the cost of the minimum set cover problem.
Therefore, the minimum set cover problem corresponding to self-information configures $n$ to $\lceil\frac{1}{p(E_{\alpha})}\rceil$, 
which means that there are $\lceil\frac{1}{p(E_{\alpha})}\rceil$ elements in the universe $U_{\alpha}$ 
of minimum set cover problem by definition of cardinality.
Now, we will see the properties of each element of universe $U_{\alpha}$.

Lebesgue integral for non-negative measure is defined as equation (28) and equation (29). 
\begin{equation}
\label{APP_D_def:lebesgue_integral}
\int f d\mu  = \sup_s\{\int_k a_k \cdot \mu(A_k)\}\ \ (0 \le h \le  f)
\end{equation}
\begin{equation}
    \label{def:lebesgue_measurable_set}
    \begin{split}
        A_k = \{ x | s(x)=a_k \}
    \end{split}
\end{equation}

Based on equation (27), we can match $a_k$ to $p(E_{\alpha})$ and $\mu(A_k)$ to $\mu(u)$ 
whose value is $1$ in equation (27).
\begin{align}
\label{connect_entropy_msc}
\begin{split}
\sup\{\sum_{n=1}^{\lceil{1 \over p(E_\alpha)}\rceil}p(E_\alpha) \cdot 1\} 
&= \sup\{\sum_{n=1}^{\lceil{1 \over p(E_\alpha)}\rceil}p(E_\alpha)\cdot \mu(u)\}
\end{split}
\end{align}
With the correspondence of equation (29) and (30), we can define $U_{\alpha}$ as follows. 
\begin{equation}
    \label{lebesgue_set_for_entropy}
    \begin{split}
        U_{\alpha} = \{ u | s(u)=p({E_{\alpha}}) \}
    \end{split}
\end{equation}

Now, we prove that there is another minimum set cover problem as predecessor linked with the problem which we have treated until now.
Since probability is Lebesgue measure and $p(r)$ is non-negative measurable function as unit probability, 
we can write $p({E_{\alpha}})$ as follows based on equation (28).
\begin{equation}
    \label{extension_msc}
    \begin{split}
        p({E_{\alpha}}) &= \int_e p(e) d\chi(e \in u)\  \textrm{s.t.}\ p(u)=p({E_{\alpha}})\\
                       &=\sup\{\int_{e}s(e)d\chi(e \in u) \}\ (0 \le s(e) \le p(e)) \\
                       &=\min\max\{ \int_{e}s(e) d\chi(e \in u) \} 
    \end{split}
\end{equation}
Since all outputs of $\max\{ \int_{e}s(e) d\chi(e \in u) \}$ should be bigger than 
or equal to $p({E_{\alpha}})$, 
we can rewrite $p(E_{\alpha})$ to equation (32) with inequality condition.
\begin{equation}
    \label{refined_extension_msc}
    \begin{split}
        p(E_{\alpha}) &= \min_{u \subset U}\{ \int_{e}s(e) d\chi(e \in u) \} \\
        &\textrm{s.t.}\ \ p({E_{\alpha}}) \le p(u)
    \end{split}
\end{equation}
We rewrite equation (33) to equation (34) in terms of arguments
and interpret $U_{\alpha}$ as the collection of $u^{\ast}$s.
\begin{equation}
    \label{element_predec_msc}
    \begin{split}
        u^{\ast} &= \arg\min_{u \subset U}\{ \int_{e}s(e) d\chi(e \in u) \} \\
        &\textrm{s.t.}\ \ p({E_{\alpha}}) \le p(u)
    \end{split}
\end{equation}

Based on this reformulation and the definition of Lebesgue integral, we can derive that each element of $U_{\alpha}$ is a set
since argument of min operation should be a set from the definition of Lebesgue integral.
To summarize, $U_\alpha$ is a collection of $\lceil \frac{1}{p(E_{\alpha})} \rceil$ arguments 
which are sets as solutions of equation (34).
We can also interpret $\min$ as greedy decision based on $\int_{e}s(e) d\chi(e \in u)$ and,
$U_{\alpha}$ is constructed with this greedy algorithm under the constraints that iteration should be repeated  
until unique $\lceil \frac{1}{p({E_{\alpha}})} \rceil$ elements are selected.

We think about the base case that $p(e)=({1 \over 2})^k$ and $p({E_{\alpha}})=(\frac{1}{2})^m\ (m \le k)$.
Since the number of elements in $U_{\alpha}$ is $2^m$ and probability of each element is $(\frac{1}{2})^m$, 
summation of all probabilities in $U_{\alpha}$ becomes 1. 
Therefore, if all elements are disjoint to each other, then $U_{\alpha}$ becomes set cover 
for given universe $U$ as a sample space which has all possible outcomes as elements.
The candidates of predecessor minimum set cover problem are subsets of given universe $U$ 
whose probability is bigger than or equal to $p_{E_{\alpha}}$ 
and cost function is simple summation across probabilities of all elements for each subset. 
Then, this $U_{\alpha}$ becomes greedy solution of the minimum set cover problem, 
since 1 becomes minimum cost of this problem for sample space.
In conclusion, universe of successor minimum set cover problem is solution of another minimum set cover problem 
whose cost function is simple summation of probabilities for all elements in that subset
since $U_{\alpha}$ is also the universe of successor minimum set cover problem.  
\end{proof}

Now, we extend the relationship to general probabilities ($p({E_{\alpha}})$) 
and general cardinalities($\ln\frac{1}{p({E_{\alpha}})}$).
\begin{proof}
Let's return to equation (28) to leverage the property of simple function. 
As a result of equation (28), we can decide $p({E_{\alpha}})$ as the simple function of Lebesgue integral 
$\int p({E_{\alpha}})d{1\over p({E_{\alpha}})}$.
According to (\cite{KwonY12}), every real function can be approximated with simple function 
whose value is $\frac{a}{2^n}$.
Since the probability of each element in $U$ is $\frac{1}{2^m}$, 
the probability of each element $u$ in $U_{\alpha}$ which is same as $p({E_{\alpha}})$ 
is the form of $\frac{a}{2^m}$ since $u$ is the set of elements in $U$.
Therefore, the greedy process can approximate every real function with the elements $e$ in $U$.
Eventually, the relationship between self-information and tight upper bound of minimum set cover 
constructed by greedy algorithm approximately holds true for all real probabilities.
In addition, it is well known that $\ln\lceil\frac{1}{p}\rceil$ becomes $\ln\frac{1}{p}$ in probability
by Asymptotic Equipartition Property(AEP) in information theory.
In conclusion, the relationship in base case also holds true for the case of general probability and cardinality 
in probability.
\end{proof}

As the last step of this proof, we show the weak duality of minimum entropy problem and the chain of 
two minimum set cover problem.

\begin{proof}
For all events $\alpha$, the following inequality about cost of minimum set cover 
($c_{E_\alpha}$) always holds true. 
\begin{equation}
\label{cost_up_bound_event}
c_{E_\alpha} \le \ln {1 \over p({E_\alpha})}
\end{equation}
Since $p_{E_\alpha}$ is positive, the multiplication of $p({E_{\alpha}})$ does not 
change the direction of inequality.
\begin{equation}
\label{weighted_cost_up_bound_event}
p({E_\alpha})c_{E_\alpha} \le p({E_\alpha})\ln {1 \over p({E_\alpha})} \\
\end{equation}
We can get the following inequality by summation across all $\alpha$s.
\begin{equation}
\label{cost_up_bound_event_space}
\sum_{\alpha} p({E_\alpha})c_{E_\alpha} \le \sum_{\alpha}p({E_\alpha})\ln {1 \over p({E_\alpha})} \\
\end{equation}
\begin{equation}
\label{cost_up_bound_expectation}
\mathbb{E}_{\alpha}[c_{E_\alpha}] \le \mathbb{E}_{\alpha}[\ln {1 \over p({E_\alpha})}]
\end{equation}
The RHS is entropy since definition of entropy is the expectation of self-information across all events. 
The LHS is expectation of cost for minimum set cover constructed by greedy algorithm. 
Therefore, the problem to minimize entropy is dual form of problem to maximize 
the expectation of cost for minimum set cover by weak duality theorem. 
Then we can extend the result to general probability and cardinalities same as the proof of Theorem 2.1.
\end{proof}

\subsection{Proof for Theorem 2.2}
Now, we prove by construction that the chain of two minimum set cover problems clusters elements of the universe 
for predecessor minimum set cover problem.
Likewise Theorem 2.1, we first prove Theorem 2.2 for the base case whose price for each element is ${1 \over 2^k}$ 
and probability of given event is ${1 \over 2^m}$ ($k \geq m$).
Then, the greedy algorithm for predecessor selects candidate with simple summation 
across the probabilities of elements until $U$ is covered. 
\begin{proof} We start with predecessor minimum set cover problem.  

\textbf{Predecessor Minimum Set Cover Problem: } If the universe for predecessor is $U=\{e_1,\ldots,e_{|U|}\}$, 
then the set of candidates becomes power set of $U$ whose element is bigger than or equal to probability corresponding to specific event $E_\alpha$. 
Then, the greedy algorithm for predecessor selects candidate with simple summation across the probabilities of elements 
for that candidate as subset of $U$ until $U$ is covered. 
We call this result as $U_\alpha$ which consists of subset whose price is ${1 \over 2^m}$ and $|U_\alpha|$ becomes $2^{k-m}$.
In this base case, each element of $|U_{\alpha}|$ is disjoint to each other since it minimizes the setcover's cost 
(This property approximately extends to general case because every real number can be approximated with ${1 \over 2^k}$).

Next we continues to successor minimum set cover problem.

\textbf{Successor Minimum Set Cover Problem: }
This $U_\alpha$ becomes universe of successor minimum set cover problem.
In this time, the candidates becomes power set of $U_\alpha$ with no constraints.
Then, greedy algorithm for successor selects candidate with optimal measure to each subset for the minimum set cover problem.
In this time, greedy algorithm selects the candidate which includes only one element of $U_{\alpha}$ due to the disjointness of each element.
Then the number of selections becomes $2^m$, which makes supremum of the cost about successor’s set cover $\ln 2^m$ 
(For the base case, we don’t have to think about the supremum. However, we can easily extend the result of base case to general case by doing this).
We define the result as $S_{\alpha}$.
Now, we think about the probability of each element in $S_{\alpha}$.
Since each element of $S_{\alpha}$ consists of the only subset whose has unique element of $U_{\alpha}$, 
the probability of each element in $S_{\alpha}$ is same as the probability of each element in $U_{\alpha}$ 
(This relationship is valid due to the definition of Lebesgue integral formulated as simple function).
Therefore, the probability of each element in $S_{\alpha}$ becomes ${1 \over 2^m}$ which is same as the given probability of event $\alpha$.

For the last, we think about the combination of results of the chain of the minimum set cover problems. 

\textbf{Combination: }
Now, we have constructions for all events $\alpha$.
Then, we select only one element of $S_{\alpha}$ for each $\alpha$ due to the property 
that all elements of $U_{\alpha}$ is equal to each other (This corresponds to each calculation of $p_{\alpha}\ln{1 \over p_\alpha}$). 
This condition usually meets in classification setting since there is implicit assumption for the disjointness across all events.
With disjointness of events,  we can say that minimum entropy problem according to the following definition according to (\cite{Soni12}).

\begin{definition}
\label{def:clustering}
Clustering  is the process of partitioning a dataset into subsets with some selected distance measure
\end{definition}
\end{proof}

\subsection{Example of Theorem 2.2}
We provide the example about the way how the chain of two minimum set cover problems works
and explains about the reason why it is important to reflect on the relationships betweeen disentangled representations.
We think about the situation that find the codebook about characters ($a$, $b$ and $c$)
and the probability of each character is $p(a)={1 \over 2}$ and $p(b)=p(c)={1 \over 4}$.
According to Theorem 3.2, the chain of two minimum set cover problems works for each character.  
Regardless of the characters, the universe of predecessor becomes $\{00, 01, 10, 11\}$.
However, the candidates becomes different due to the discrepancies of probabilities for each character.
These discrepancies make all following matters different. 

\textbf{Clustering for $a$: }
The set of candidates for a becomes 
$\{\{00, 01\}, \{00, 10\}, \{00, 11\}, ..., \{00, 01, 10, 11\}\}$.
The greedy algorithm of predecessor only selects one of the candidates based on the scale of probability 
for each set. 
Since all probabilities are bigger than ${1 \over 2}$ and there are more than 2 candidates set of candidates,
greedy algorithm selects only candidates whose probabilities are ${1 \over 2}$ 
among $\{\{00, 01\}, \{00, 10\}, \{00, 11\}, \{01, 10\}, \{01, 11\}, \{10, 11\}$).
We assume that greedy algorithm selects $\{00, 10\}$ at first and $\{01, 11\}$ for the next, like Figure 5 
(This is possible since greedy algorithm does not consider the relationship between elements of $U$ by itself).
Then, the set cover $U_a$ ($\{\{00, 10\},\{01, 11\}\}$) becomes new universe of successor minimum set cover problems. 
In the case of successor, the candidates becomes power set of $U_{\alpha}$ excluding $\emptyset$.
Then, the greedy algorithm which considers about the overlaps selects the subset 
which consists of the only element in $U_{\alpha}$ like Figure 5 for each iteration. 
Then, the output setcover of successor becomes $\{\{\{00,10\}\},\{\{01,11\}\}\}$ 
due to the disjointness of each element to the other elements in $U_a$.

\textbf{Clustering for $b$ and $c$: }
The same process also proceeds in the case of characters $b$ and $c$ for different probabilities and candidates. 
For the character $b$ and $c$, the candidates of predecessor becomes $\{\{00\}, \{01\}, \{10\}, \{11\},....,\{00,01,10,11\}\}$
$(P(\{00,01,10,11\})\backslash \{ \emptyset \})$ since probabilities for $b$ and $c$ are ${1 \over 4}$ and 
the probability for each candidate should be bigger than or equal to the probability of corresponding character.
Likewise character $a$, the greedy algorithm of predecessor selects the candidates whose probabilities are ${1 \over 4}$.
We also assume that greedy algorithm selects $\{00\}, \{01\}, \{10\}$ and $\{11\}$ for each iteration,
which leads to set cover $\{\{00\}, \{01\}, \{10\}, \{11\}\}$ as $U_{b}$ and $U_{c}$
These set covers becomes universe again for the next minimum set cover problem for character $b$ and $c$.
The candidates for successor become the power set of $U_b$ and $U_c$ without $\emptyset$. 
The successor's greedy algorithm also selects the subset of only one element in $U_b$ and $U_c$
in consideration with overlaps for each iteration like Figure 6 and Figure 7.
Then, the output setcover of successor becomes $\{\{\{00\}\},\{\{01\}\},\{\{10\}\},\{\{11\}\}\}$

\textbf{Selection of Clusters for $a$, $b$ and $c$: }
We have the following set covers for $a$, $b$, and $c$.
\begin{align}
\label{cluster_result}
    S(a)&=\{\{\{00,10\}\},\{\{01,11\}\}\}\\
    S(b)&=\{\{\{00\}\},\{\{01\}\},\{\{10\}\},\{\{11\}\}\}\\
    S(c)&=\{\{\{00\}\},\{\{01\}\},\{\{10\}\},\{\{11\}\}\}
\end{align}
Now, we select only one element in set cover for each character 
under the constraint that each element is disjoint to each other. 
For example, we select $\{\{00,10\}\}$ for a, $\{\{01\}\}$ for b and $\{\{11\}\}$ for c
like Figure 8.  



\textbf{Analysis: }
Generally, an optimal coding book for bit strings is designed under the condition 
that the distance between bit strings and the context that bits are read 
from the leftmost position of the bit string are given.
As a result, the bit string 00 is closest to 01, followed by 10, and finally 11.
In contrast, the provided example corresponds to a situation 
to design the optimal coding book, given only minimum entropy.
As a result, the coding book constructed in the example differs from the optimal coding book.
In other words, an optimal coding book cannot be designed based solely on minimum entropy
(additional information about the relationships between bit strings is required to construct the optimal coding book).

In training a neural network, no information is provided about the relationships between their disentangled representations unlike designing an optimal coding book for bit strings.
While supervised signals generally provide information about the relationships 
between samples or between samples and labels, 
the absence of direct labels for the disentangled representations means that their relationships remain unknown.
This remains true even in InfoNCE, which has become the new standard.
In other words, without additional information about the relationships between disentangled representations, 
InfoNCE cannot leverage these relationships, even though it can utilize the relationships between samples.
However, the floating boundary problem causes the relationships between disentangled representations 
to vary based on the context (e.g., decoding from the front or back), 
making it difficult to provide direct supervised signals for these relationships.
To address this issues, we have developed URECA.

\newpage
\begin{figure}[!h]
\centering
\includegraphics[width=0.95\columnwidth]{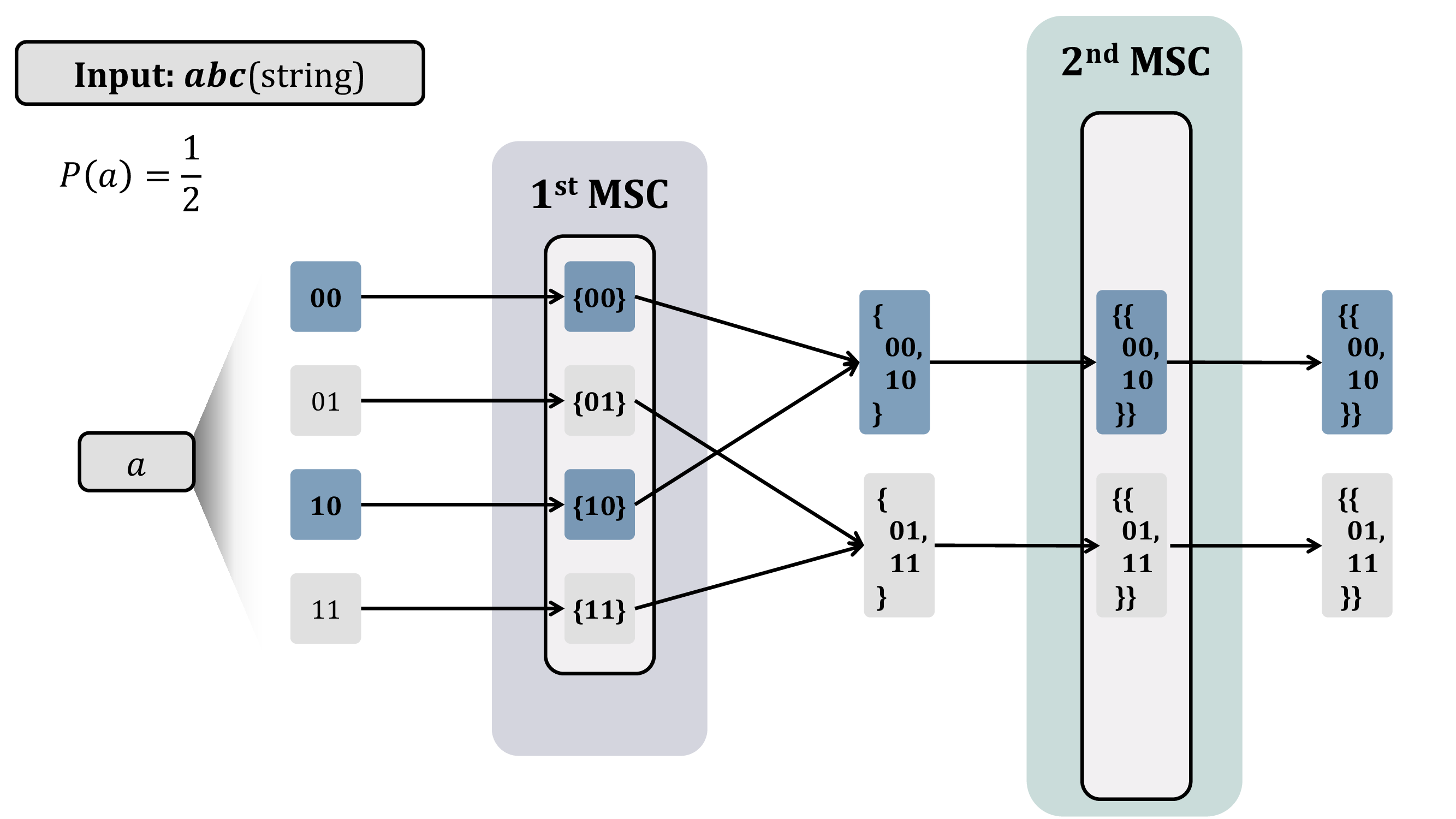} 
\caption{Clustering behind Minimum Entropy Problem for a}
\label{fig_ex1}
\end{figure}

\begin{figure}[!]
\centering
\includegraphics[width=0.95\columnwidth]{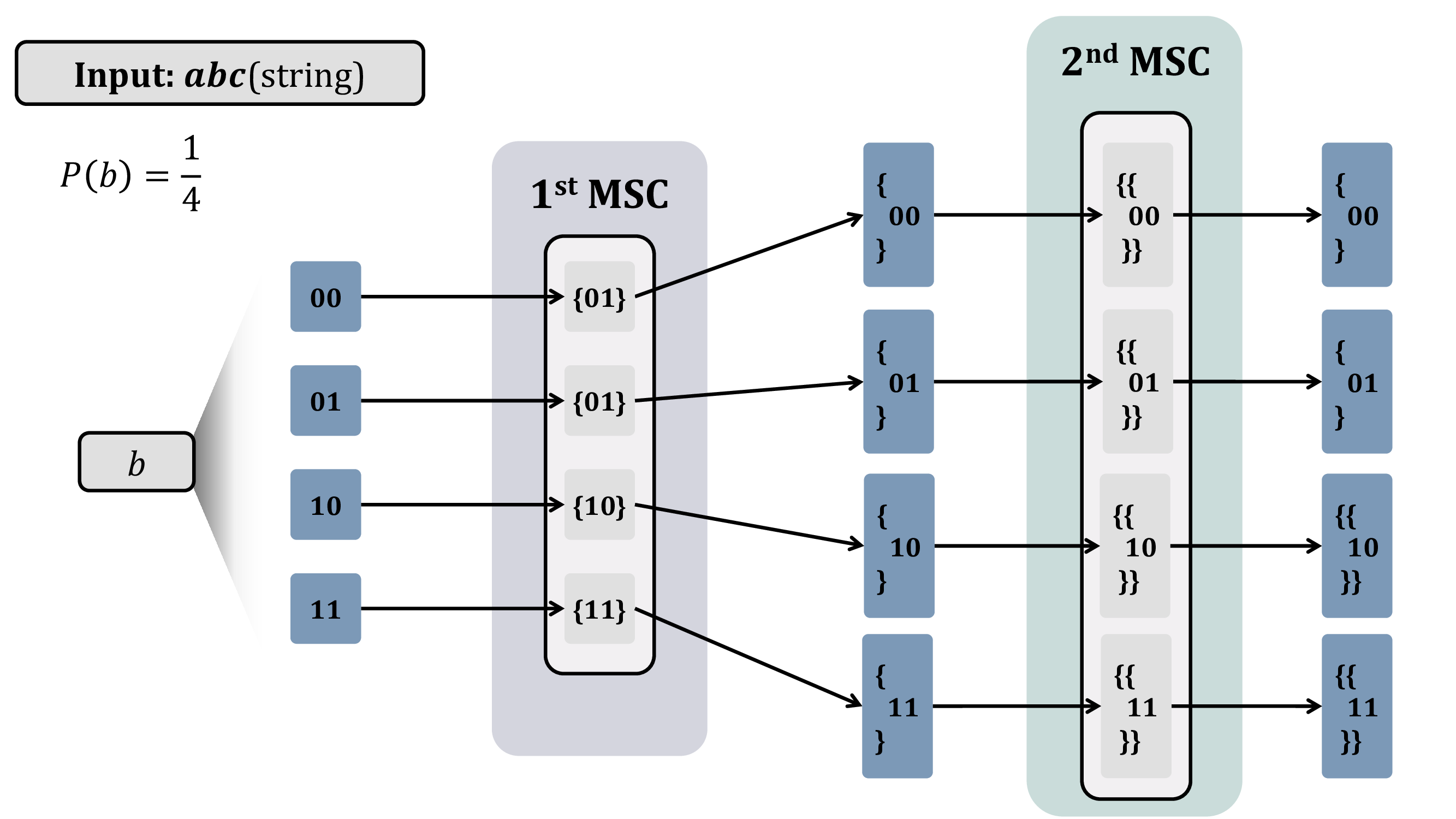} 
\caption{Clustering behind Minimum Entropy Problem for b}
\label{fig_ex2}
\end{figure}

\newpage
\begin{figure}[!t]
\centering
\includegraphics[width=0.994\columnwidth]{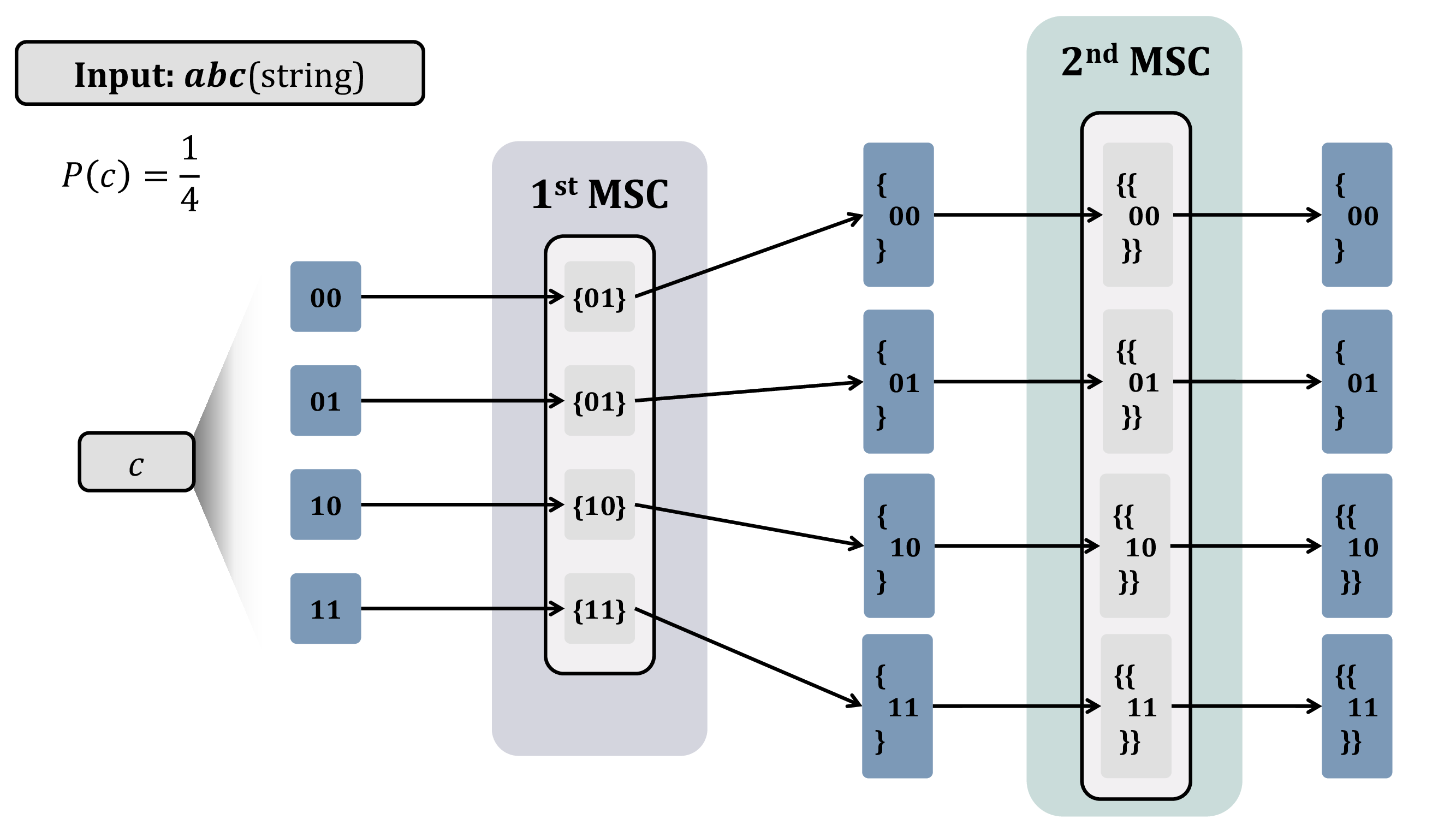} 
\caption{Clustering behind Minimum Entropy Problem for c}
\label{fig_ex3}
\end{figure}

\begin{figure}[!]
\centering
\includegraphics[width=0.998\columnwidth]{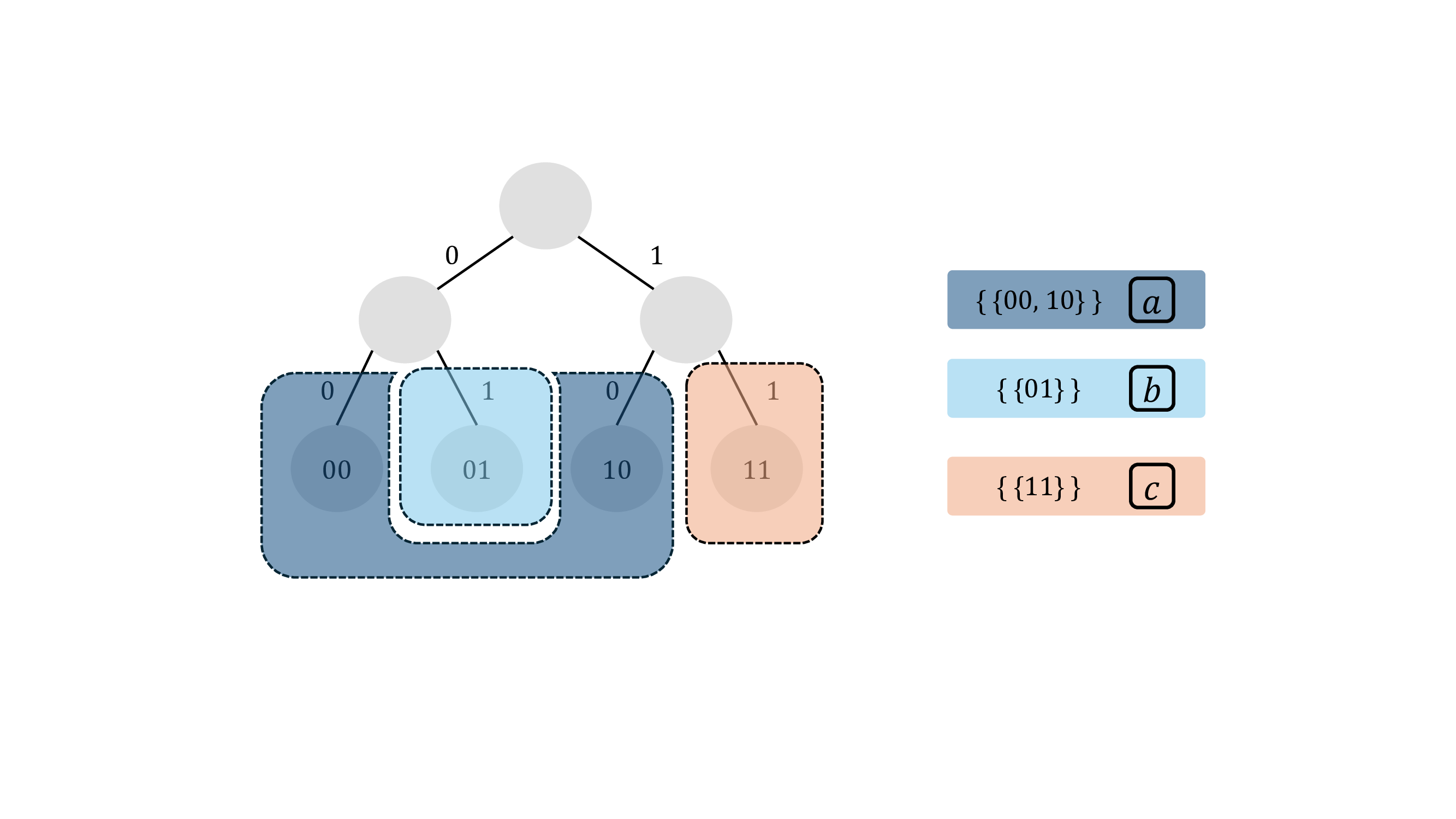} 
\caption{Cluster Selection for a, b and c }
\label{fig_ex4}
\end{figure}

\newpage
\subsection{Proofs for Theorem 3.1 and Corollary 3.2}

To prove Theorem 3.1, we first define Usually Positive Limit which is widely used in our proofs. 
\begin{definition}
\label{def:up-limit}
For an sequence $a_n$, if there exists $N \in \mathbb{N}$ such that 
for $\forall n \geq N$, $|a_n-\alpha| < \epsilon$ ($\forall \epsilon \in \mathbb{R}^+$), 
then $a_n$ converges to constant $\alpha$ in terms of Universally Positive Limit (UP-Limit).
The notation is as follows. 
\begin{align}
    \lim^{UP}_{n \rightarrow \infty}a_n=\alpha  
\end{align}
\end{definition}

\begin{lemma}
\label{lem:adj_convergence}
if the UP-Limit of $p(C^t_{i,j})$ converges to ${1 \over |J|}$ (J is the set of cluster indexes ),
then the UP-Limit of $evid(C^{t+1}_{i,j})-evid(C^{t+1}_{a,b})$ converges to zero
($\forall i,j, a, b \in J$). 

\begin{align}
    \begin{split}
        \lim^{UP}_{t \rightarrow \infty}p(C^{t}_{i,j})={1 \over |J|} \implies
        \lim^{UP}_{t \rightarrow \infty}(evid(C^{t+1}_{i,j}-evid(C^{t}_{a,b})))=0
        \ \ (\forall i,j,a,b \in J)
    \end{split}
\end{align}
\end{lemma}
\begin{proof}
By definition of UP-Limit, the following statements holds true for the sequence $p(C_{t,j})\ (\forall j \in J)$.
There exists $T \in \mathbb{N}$ 
such that $|p(C_{t,j})-{1 \over |J|}| < \epsilon\ (\forall \epsilon \in \mathbb{R}^+)$.
\begin{align}
    \begin{split}
        &\exists T \in \mathbb{N}\\
        s.t. |p(&C^t_{i,j})-{1 \over |J|}| < \epsilon \\
        (\forall t &\geq T,\ \forall \epsilon \in \mathbb{R}^+) 
    \end{split}
\end{align}

Based on the equation (44), we can derive the following inequalities about $evid(C^t_{i,j})$. 
\begin{align}
           &{1 \over |J|} - \epsilon < p(C^t_{i,j})  <  {1 \over |J|} +\epsilon (\forall \epsilon \in \mathbb{R}^+)\\
&\implies   \ln{{1 \over |J|} - \epsilon \over p(y)} < \ln{p(C^t_{i,j})\over p(y)} < \ln{{1 \over |J|} + \epsilon \over p(y)}\\
&\implies   \ln{{1 \over |J|} - \epsilon \over p(y)} < evid(C^t_{i,j}) < \ln{{1 \over |J|} + \epsilon \over p(y)}\\
&
\end{align}
With these inequalities, we can obtain the following relationship about the difference $evid(C^{t+1}_{i,j})$ and $evid(C^t_{a,b})$.
\begin{align}
    -\ln{{1\over|J|} + \epsilon \over {1 \over |J|} - \epsilon} < evid(C^{t+1}_{i,j}) - evid(C^t_{a,b}) 
    < \ln{{1\over|J|}+\epsilon \over {1 \over |J|} -\epsilon}
\end{align}

We can derive following statement from equation (49) when, 
$\delta = \ln{{1 \over |J|}+\epsilon \over {1 \over |J|} -\epsilon} \ \  (\forall \epsilon \in \mathbb(R)^{+})$. 
\begin{align}
    \exists T &\in \mathbb{N}\\
    s.t.\ \ |evid(C^{t+1}_{i,j})-e&vid(C^t_{a,b})| < \delta \ \ (\forall \delta \in \mathbb(R)^+)
\end{align}
By the definition of universally positive limit, $evid(C^{t+1}_{i,j})-evid(C^{t}_{a,b})$ converges to zero as $t$ goes to $infty$
\begin{align}
    \lim_{t\rightarrow \infty}^{UP}(evid(C^{t+1}_{i,j})-evid(C^{t}_{a,b}))=0
\end{align}
Therefore, the only condition that the universally positive convergence of $P(C^t_{i,j})$ to ${1 \over |J|}$
implies that the sequence of $P(C^t_{i,j})$ is Uniform Cauchy.
\end{proof}

\begin{lemma}
\label{lem:adj_lip_continuous}
If the UP-Limit of $evid(C^{t+1}_{i,j})-evid(C^t_{a,b})$ converges to 0 ($\forall i,j,a,b \in J$),
then $evid(C^t_{i,j})$ is $\ln{{1 \over |J|}+\epsilon \over {1 \over |J|}-\epsilon}$-Lipschitz Continuous
($\forall \epsilon \in \mathbb{R}^+$).
\end{lemma}
\begin{proof}
Based on equation (49), we can extend the relationship between difference of evidences from adjacent timesteps to general timesteps
for all $i,j,a,b \in J$.  
\begin{align}
    \begin{split}
        -1\cdot(\ln{{1 \over |J|}+\epsilon \over {1 \over |J|}-\epsilon}) 
        < evid(C^{t+1}_{i,j}) &- evid(C^{t}_{a,b}) < 1\cdot(\ln{{1 \over |J|} + \epsilon \over {1 \over |J|}-\epsilon})\\
        -2\cdot(\ln{{1 \over |J|}+\epsilon \over {1 \over |J|}-\epsilon}) 
        < evid(C^{t+2}_{i,j}) &- evid(C^{t}_{a,b}) < 2\cdot(\ln{{1 \over |J|} + \epsilon \over {1 \over |J|}-\epsilon})\\
                              &\dots\\
        -L\cdot(\ln{{1 \over |J|}+\epsilon \over {1 \over |J|}-\epsilon}) 
        < evid(C^{t+L}_{i,j}) &- evid(C^{t}_{a,b}) < L\cdot(\ln{{1 \over |J|} + \epsilon \over {1 \over |J|}-\epsilon})\\        
    \end{split}
\end{align}
We can rewrite the last inequality in equation (53) as follows. 
\begin{align}
    |{evid(C^{t+L}_{i,j}) - evid(C^t_{a,b}) \over (t+L) - (t)}| < \ln {{1 \over |J|} +\epsilon \over {1 \over |J|} - \epsilon}
\end{align}
In conclusion, $evid(C^{t}_{i,j})$ is $\ln{{1 \over |J|}+\epsilon \over {1 \over |J|}-\epsilon}$ Lipschitz-continuous 
by the definition of Lipschitz-continuous~(\cite{KavoshDM18}).  
\end{proof}

However, the $\ln{{1 \over |J|}+\epsilon \over {1 \over |J|}-\epsilon}$ Lipschitz-continuity does not assure the uniform convergence. 
So we dervie the following lemma.

\begin{lemma}
\label{lem:uniform_cauchy}
if the UP-Limit of $P(C^t_{i,j})$ converges to ${1 \over |J|}$ ($\forall i,j,a,b \in J$) 
and $evid(C^t_{i,j})$ is $\alpha\cdot\ln{{1\over|J|}+\epsilon \over {1\over|J|}-\epsilon}$-Lipschitz continuous 
($\forall \alpha \in [0,{{1 \over |J|}-\epsilon \over {1 \over |J|}+\epsilon})$),
then $evid(C^t_{i,j})$ is uniform cauchy sequence of $t$.
\end{lemma}
\begin{proof}
By Lemma D.3, the following inequality holds true. 
\begin{align}
    -L\cdot(\ln{{1 \over |J|}+\epsilon \over {1 \over |J|}-\epsilon})
    < evid(C^{t+L}_{i,j})-evid(C^{t}_{a,b})
    <L\cdot(\ln{{1 \over |J|}+\epsilon \over {1 \over |J|}-\epsilon})\ \ (\forall i, j, a, b \in J)
\end{align}
In general, the following inequality holds true. 
\begin{align}
    {d \over dx}(\ln x)={1 \over x} < 1 = {d \over dL}(L)\ (x > 1)
\end{align}
Equation (56) implies $L\cdot \ln{{1 \over |J|}+\epsilon \over {1 \over |J|}-\epsilon}$ diverges to positive infinity.
\begin{align}
    \lim_{L\rightarrow \infty, \lim \epsilon \rightarrow 0} (L\cdot\ln{{1 \over |J|}+\epsilon \over {1 \over |J|}-\epsilon})=\infty 
\end{align}

However, if we multiply $\alpha \in [0, {{1 \over |J|}-\epsilon \over {1 \over |J|}+\epsilon}]$ 
with $L\cdot \ln{{1 \over |J|}+\epsilon \over {1 \over |J|}-\epsilon}$, 
then $(\alpha L)\cdot \ln{{1\over|J|}+\epsilon \over {1 \over ||J}-\epsilon}$ converges to 0 for universally positive limit. 
\begin{align}
    \lim^{UP}_{L \rightarrow \infty, \epsilon \rightarrow 0}|(\alpha L)
    \cdot \ln{{1 \over |J|} + \epsilon \over {1 \over |J|} - \epsilon}| = 0 
\end{align}
This is because the following relationship.
\begin{align}
    \begin{split}
       &{d \over d x}(\ln x) = |{1 \over x}| > \alpha = ({d \over dL}(\alpha \cdot L)) \\
        (\forall &x \in [0, {{1 \over |J|} +\epsilon \over {1 \over |J|}-\epsilon}),
        \forall \alpha \in [0, {{1 \over |J|}-\epsilon \over {1 \over |J|}+\epsilon}))
    \end{split}
\end{align}
Therefore, the conditions of universally positive convergence of $P(C^t_{i,j})$ and
$(\alpha L)\cdot \ln{{1 \over |J|}+\epsilon \over {1 \over |J|}-\epsilon}$-continuous imply that 
$|evid(C^{t+L}_{i,j})-evid(C^t_{a,b})| < \alpha\ (\forall \alpha \in \mathbb{R}^+)$.
By the definition of uniform cauchy condition, the sequence $evid(C^t_{i,j})$ is uniform cauchy. 

\end{proof}

Based on these Lemmas, we prove the Theorem 3.1 that $evid(C^t_{i,j})$ converges uniformly.
We first show the pointwise convergence of $evid(C^t_{i,j})$.
\begin{proof}
By Lemma D.4, $evid(C^t_{i,j})$ is uniform cauchy under the conditions of Theorem 3.1. 
The, for each $i, j \in J$, the real sequence $evid(C^t_{i,j})$ is Cauchy, 
so it converges by completeness of $mathbb{R}$.
We define $evid(C_{i,j})$ as follows, and then $evid(C^t_{i,j}) \rightarrow evid(C_{i,j})$ pointwise.
\begin{align}
    evid(C_{i,j}) = \lim_{t \rightarrow \infty} evid(C^{t}_{i,j})
\end{align}

Next, we prove the uniform convergence of $evid(C^t_{i,j})$.
If $\gamma > 0$, then we can choose $T \in \mathbb{N}$ (depending only on $\gamma$) such that 
$|evid(C^m_{i,j})-evid(C^n_{a,b})| < {\gamma \over 2}\ (for\ \forall i,j,a,b \in J and m, n > T)$.
In this case, we have the following inequality. 
\begin{align}
    \begin{split}
        |evid(C^m_{i,j})-evid(C_{i,j})| 
        &\leq |evid(C^m_{i,j})-evid(C^n_{a,b})| + |evid(C^n_{a,b})-evid(C_{i,j})|\\
        &< {\gamma \over 2} + |evid(C^n_{a,b})-evid(C_{i,j})|
    \end{split}
\end{align}
Since $evid(C^n_{a,b})\rightarrow evid(C_{a,b})$ as $m\rightarrow \infty$, 
we can choose $m > T$ for the following inequality. 
\begin{align}
    |evid(C^n_{a,b})-evid(C_{a,b})| < {\gamma \over 2}\ (\forall a, b \in J)
\end{align}
Since there is no constraint about $a, b$ of $C_{a,b}$ for the following condition,
\begin{align}
    \lim^{UP}_{t \rightarrow \infty}P(C^t_{a,b})={1 \over |J|} (\forall a,b \in J)
\end{align}
, we can rewrite the equation (62) as follows. 
\begin{align}
    |evid(C^n_{a,b})-evid(C_{i,j})| < {\gamma \over 2}\ (\forall i,j,a,b \in J)
\end{align}
Equation (61) derives the following,
\begin{align}
    |evid(C^{m}_{i,j})-evid(C_{i,j})| < \gamma\ (\forall j \in J)    
\end{align}
Equation (65) means that $evid(C^{m}_{i,j})$ uniformly converges to $evid(C_{i,j})$.

\end{proof}

At last, we prove the Corollary 3.2 as a refined version of Theorem 3.1. 
\begin{proof}
As we have done so far, We define $evid(C^{t+1}_{i,j})$ and $evid(C^t_{a,b})$ as follows.
\begin{align}
    evid(C^{t+1}_{i,j}) &= \ln{p(C^{t+1}_{i,j}) \over p(y)}\\
    evid(C^t_{a,b}) &= \ln{p(C^t_{a,b}) \over p(y)}
\end{align}
For $evid(C^{t+1}_{i,j})$ and $evid(C^t_{a,b})$, the following holds true by Jenson's Inequality. 
\begin{align}
    \begin{split}
        evid(C^{t+1}_{i,j}) &= \ln{p(C^{t+1}_{i,j}) \over p(y)}\\
                        &= \ln E_{(a,b) \sim p(C^{t+1}_{i,j}|C^t_{a,b})}[{p(C^t_{a,b}) \over p(y)}]\\
                        &\geq E_{(a,b) \sim p(C^{t+1}_{i,j}|C^{t}_{a,b})}[\ln{p(C^t_{a,b})\over p(y)}]
    \end{split}
\end{align}

The equality of equation (68) holds true when all arguments of convex function are same 
$(\forall a, b \in J)$. 
However, if $C^{t+1}_{i,j}$ and $C^{t}_{a,b}$ are mutually exclusive, then 
$p(C^{t+1}_{i,j}|C^{t}_{a,b})$ becomes zero. 
In this case, the term $p(C^{t+1}_{i,j}|C^{t}_{a,b})\ln {p(C^{t}_{a,b}) \over p(y)}$
is ceared out in the convex combination. 
Since these kinds of clusters do not impact on the equality condition after some threshold,
we conclude that we can distill the entire index set $J$ for all clusters
to $K$ like Corollary 3.2.
The definition of $K$ is as follows. 
\begin{align}
    K = \{j | \lim^{UP}_{t \rightarrow \infty} p_t(C^{t+1}_{i,j}|C^{t}_{a,b})=0 ,j \in J\} 
\end{align}
Since the only difference between Theorem 3.1 and Corollary 3.2 is replacement $J$ with $K$, 
the Corollary 3.2 also holds true. 
\end{proof}
\section{Experiments}
\subsection{Taxonomy}

Table~\ref{tbl:taxonomy} presents a taxonomy of shift settings in model evaluation,
categorizing different experimental scenarios based on the presence of task, query, and code shifts.
The columns represent the addition of a new shift, while the rows indicate the existing shift settings.
The entries in the table denote whether a setting is feasible, covered in the extensive experiments including ablation studies.
For instance, when you see the ``\textbf{[No Shift]}'' row with ``\textbf{+ Query Shift}'' column,
CoCoSoDA is evaluated on CSN, without query shift, and CoSQA, with query shift.
This demonstrates the query shift ablation to examine how query variations affect the performance.
``\textbf{IMPOSSIBLE}'' denotes the settings that are infeasible,
such as applying two independent task shifts.
``X'' denotes settings that are theoretically possible but not included in the current experiments.

\begin{table}[h]
    \resizebox{\linewidth}{!}{
        \setlength{\tabcolsep}{6pt} 
        \begin{tabular}{p{4cm} p{3.5cm} p{3.5cm} p{3.5cm}}
            \toprule
            \textbf{Shift Type} & \textbf{+ Task Shift} & \textbf{+ Query Shift} & \textbf{+ Code Shift} \\
            \midrule
            \textbf{[No Shift]} & [CoCoSoDA] vs [UniXCoder] (CSN) & CoCoSoDA for [CSN] vs [CoSQA] & X \\\hline
            \textbf{Task Shift} & \textbf{IMPOSSIBLE} & UniXCoder for [CSN] vs [CoSQA] & [CodeT5+] vs [UniXCoder] (CSN) \\\hline
            \textbf{Query Shift}& [CoCoSoDA] vs [UniXCoder] (CoSQA) & \textbf{IMPOSSIBLE} & X \\\hline
            \textbf{Code Shift} & X & X & \textbf{IMPOSSIBLE} \\\hline
            \textbf{(Task Shift, Query Shift)} & \textbf{IMPOSSIBLE} & \textbf{IMPOSSIBLE} & [CodeT5+] vs [UniXCoder] (CoSQA) \\\hline
            \textbf{(Task Shift, Code Shift)}  & \textbf{IMPOSSIBLE} & CodeT5+ for [CSN] vs [CoSQA] & \textbf{IMPOSSIBLE} \\\hline
            \textbf{(Query Shift, Code Shift)} & X & \textbf{IMPOSSIBLE} & \textbf{IMPOSSIBLE} \\
            \bottomrule
        \end{tabular}
    }
    \caption{The taxonomy of experiments considering shift settings, where ``\textbf{[No Shift]}'' serves as the baseline.
    Each column represents the introduction of an additional shift, enabling ablation studies to analyze its impact.}
    \label{tbl:taxonomy}
\end{table}

\subsection{Dataset}

\textbf{Statistics}
\begin{table}[h]
\centering
\begin{tabular}{lccc}
\toprule
    Dataset & Training & Validation & Test \\\toprule
    Ruby-CSN & 24,927 & 1,400 & 1,261 \\
    JavaScript-CSN & 58,025 & 3,885 & 3,291 \\
    Java-CSN & 164,923 & 5,183 & 10,955 \\
    Go-CSN & 167,288 & 7,325 & 8,122 \\ 
    PHP-CSN & 241,241 & 12,982 & 14,014 \\
    Python-CSN & 251,820 & 13,914 & 14,918 \\\midrule
    CoSQA & 19,604 & 500 & 500 \\\midrule
    CoNaLa-docprompt & 768,777 & 201 & 543 \\\bottomrule
\end{tabular}%
\caption{Dataset statistics.}
\label{datastat}
\end{table}

\textbf{CSN:}                             
CodeSearchNet~(\cite{HusainWGAB19})  includes 2.3M functions of six programming languages 
(Ruby, Javascript, Java, Go, PHP, Python) with natural language documents.
CSN is the filtered version of CodeSearchNet~(\cite{HusainWGAB19}).
The authors of (\cite{HusainWGAB19}) filter low-quality queries by handcrafted rules and 
expand 1000 candidates to the whole code corpus, which is closer to the real-life scenario. 
Since CSN is generally used in the context of code search, we evaluate URECA with this dataset.
Unlike general situation in code search, however, CodeSearchNet/CSN query as summary of corresponding code.
This property of CodeSearchNet/CSN enables us to design experiments to reveal impact of specific shift 
on performance with comparative analysis.
For example, UniXCoder was trained on CSN during the pretraining phase, 
but it was not trained with relevance between queries and code as supervised signals for code search. 
Therefore, by evaluating UniXCoder on CSN, 
we can observe how InfoNCE and URECA behaves in a task shift scenario.
Although the shift related to CodeT5p-220M may involve both query shift and code shift, 
the queries in CSN are essentially summaries of the code, making them dependent on code shift. 
Therefore, by evaluating CodeT5p-220M on CSN, we can examine how each methodology behaves in a scenario 
where code shift and task shift are combined. 
Consequently, by comparing the performance of CodeT5p-220M and UniXCoder, 
we can assess the impact of code shift on performance changes.
In this paper, we use (CoCoSoDA\cite{EnshenYWLHSDH23}) which the version to apply a data augmentation technique to UniXCoder. 
By comparing the performance of CoCoSoDA and UniXCoder on CSN, 
we can understand how the accuracy of dynamics estimation affects the performance of URECA.

\textbf{CoSQA: }
Similar to CSN, CoSQA (\cite{HuangTSG0J0D20}) is also a dataset curated for evaluating code search, 
built based on CodeSearchNet.
In the case of CoSQA, the query pool is constructed from the search logs of Microsoft's Bing search engine.
The code pool is constructed from Python code in CodeSearchNet 
that does not contain non-English documentation or special tokens (e.g., $<$img...$>$ or "http://").
Subsequently, people manually match the queries from the query pool with the code from the code pool.
This process results in a dataset that better reflects real-world scenarios compared to CodeSearchNet/CSN, 
whose queries are simply constructed as summaries of the code.
At first, We can investigate the impact of query shift alone 
by comparing the performance of CoCoSoDA for CSN and CoSQA.
Next, we can compare the performances of UniXCoder for CSN and CoSQA 
to look into the impact of query shift in the existence of task shift. 
In addition, we can compare the performances of UniXCoder and CoCoSoDa to examine the impact of task shift 
in the existence of query shift. 
We also compare the performances of CodeT5+ and UniXCoder for CoSQA to inspect the impact of code shift 
in the existence of task shift and query shift. 
Furthermore, by comparing the performance of CodeT5p-220M on CSN-Python and CoSQA, 
we can analyze the impact of query shift in scenarios where code shift and task shift are present.

\textbf{CoNaLa-docprompt: }
CoNaLa-docprompting (\cite{Zhou0XJN23}) originates from CoNaLa (\cite{PengchenBEBG18}).
CoNaLa is a dataset that recognizes the limitations of heuristic methods in dataset construction. 
It is built by curating high-quality pairs of NL queries and code, 
not only based on handcrafted features obtained through heuristic methods but also using features 
derived from probabilistic models.
CoNaLa-docprompting is a re-split version of CoNaLa, 
designed to reflect distribution shift by including one Python function in each test example 
that is not present in the training distribution.
In addition, CoNaLa-docprompting differs from CoNaLa in that it retrieves documentations 
for all Python library functions available in DevDocs and uses these to construct a documentation pool.
This structure of CoNaLa-docprompting enables the retrieval of NL documentation 
for specific functions with nl query.
In particular, CoNaLa-docprompting ensures this by removing all non-text data during its construction process.
Based on these properties of CoNaLa-docprompting, we design experiments related to GraphCodeBERT's methodology, 
including graph-guided attention and data flow augmentation, 
as well as URECA in terms of the comprehension of semantic structures in natural language.

\subsection{Baselines}
\textbf{GraphCodeBERT: }
GraphCodeBERT (\cite{GuoRLFT0ZDSFTDC21}) learns the semantic structure of code through a structure-aware pretraining task 
by employing graph-guided masked attention, which leverages the relationships 
between the specified code and data flow.
Dataflow is a graph that represents dependency relation between variables 
in which nodes represent variables and edges represent where the value of each variable comes from
in corresponding code.
Such code structure provides crucial code semantic information for code understanding.
Data flow supports the model to consider long-range dependencies induced by using the same variable or 
function in distant locations.
To incorporate the graph structure into Transformer, 
GraphCodeBERT defines a graph-guided masked attention function to filter out irrelevant signals. 
To represent dependency relation between source code tokens and nodes of the data flow, 
they define the set $E'$ of pairs of source code token and node of data flow which is identified 
from the source code token.
Then, graph-guided masked attention masks the attention score for the pairs of source code token and 
node of data flow which are not in $E'$. 
In order to learn code representations from source code and code structure, 
GraphCodeBERT introduces new structure-aware pre-training tasks with data flow and graph-guided masked attention.
One is data flow edges prediction for learning representation from code structure, 
and the other is variable-alignment across source code and dataflow for aligning representation 
between source code and code structure.
The pretraining task based on data flow and graph-guided masked attention enables GraphCodeBERT 
to learn the semantic structure of code.
URECA is similar to GraphCodeBERT in terms of the reflection of semantic structure of fragments.
Therefore, we demonstrate the effectiveness of URECA about development for semantic structures 
about fragments thorugh comparative analysis with GraphCodeBERT.

\textbf{UniXCoder: }
UniXCoder follows (\cite{GuoLDW0022}) to forward the same input with dropout mask as a positive example 
and use other representations in the same batch as negative examples.
For cross-modal generation, URECA generates its comment from the corresponding function.
Since the generation of the comment is conditioned on code, it will force the model 
to fuse semantic information from the comment into the hidden states of of the code.
UniXCoder, pretrained on CSN through Cross-Modal Generation, learns patterns of relationships 
between NL comments and code that are well-suited for generation tasks on CSN.
Therefore, by evaluating UniXCoder on the code search task using CSN, 
we can assess how effectively the patterns of relationships between NL comments and code, 
learned through generation tasks, can be adjusted for retrieval tasks. 
In other words, this evaluation allows us to measure how effectively URECA facilitates adaptation to task shifts.

\textbf{CodeT5+: }
CodeT5+ is a family of encoder-decoder based large language models (LLMs) designed 
for a broad range of code-related understanding and generation tasks.
It addresses limitations in existing code LLMs regrading architecture inflexibility 
and limited pretraining tasks with modular architectures and diverse set of pretraining tasks.
The pretraining of CodeT5+ consists of two steps. 
For the first step, they train the model with GitHub code dataset through span denoising and 
causal language modeling.
For the second step, they train the model with text-code contrastive learning and 
text-code matching with CSN dataset.
Since the bidmodally trained version of CodeT5+ is already trained for CSN with the relevance signals, 
there is no space for shifts unlike UniXCoder. 
(UniXCoder is also pretrained on CSN but there is no matching task in the process of pretraining, 
which makes it possible to train the model for task shift)
So we use the unimodally trained version of CodeT5+ in this paper. 
This unimodal version of CodeT5+ is not trained on CSN dataset unlike bimodal version.
This enables us to evaluate URECA in terms of distribution shift for codes.

\textbf{CoCoSoDA: }
CoCoSoDA (\cite{EnshenYWLHSDH23}) consists of two-stage fine-tunings.
In the first stage of fine-tuning, learning is performed using Multi-modal Contrastive Learning 
based on Soft Data Augmentation. 
In the second stage, the model is trained for the code search task through conventional contrastive learning 
on query and code pairs.
CoCoSoDA augment queries and codes with different utilizations of masking and replacement operations 
for each iteration.
There are 4 augmentation methodologies, Dynamic Masking, Dynamic Replacement and Dynamics Replace.
Except for Dynamic Masking, the others are only applicable to code data 
due to the dependence of type information which is not generally available for natural language.
CoCoSoDA trains model with inter-modal loss function and intra-modal loss function.
Inter-modal loss function is criterion to maximize the semantic similarity of 
(original query, paired code snippet (original code + augmented code)) and 
(paired query (original query + augmented query), original code) minimize the semantic similarity 
of the query and its unpaired code snippets in same mini-batch.
Intra-modal loss function is criterion to learn the better representations of queries/codes, 
where similar queries/codes have closed representations and different queries/codes have distinguishing.
This CoCoSoDA makes the model to learn more accurate dynamics of queries itself, 
codes itself and relationships between queries and codes.
Building on this, CoCoSoDA enables more accurate estimation of the dynamics used for transport in URECA. 
Based on this, we conduct experiments to evaluate how the accuracy of dynamics estimation impacts 
the effective operation of URECA.

\subsection{Metrics}
\textbf{Mean Reciprocal Rank(MRR): } 
Mean Reciprocal Rank (MRR) is a statistical measure widely used in information retrieval 
and recommendation systems to evaluate the effectiveness of a ranking algorithm. 
It is particularly helpful when there is a single correct answer or a relevant result in a ranked list.
MRR calculates the average reciprocal rank of the first relevant result across multiple queries or test cases. The reciprocal rank is the multiplicative inverse of the rank position of the first relevant result 
for a query.
The formal definition of Mean Reciprocal Rank is as follows, when $N$ is the total number of examples
and $rank_i$ is  the rank of the first relevant result for the $i$-th query. 
If there is no relevant result, the reciprocal rank is treated as 0.
\begin{align}
    MRR={1 \over N}\sum_{i=1}^N{1 \over rank_i} 
\end{align}

\textbf{Recall@k: }
Recall@k is a performance metric commonly used in information retrieval, recommendation systems, 
and machine learning to evaluate how effectively a model retrieves relevant results from a ranked list. 
It measures the proportion of relevant items successfully retrieved within the top $k$ results.
The formal definition of Recall@k is as follows, when $R_k$ is the number of relevant items in top $k$
results  and $T$ is the total number of relevant items.
\begin{align}
    Recall@k={R_k\over T}
\end{align}

\newpage
\subsection{Main Result}
In this subsection, we provide the main result of URECA for CSN-Ruby, CSN-Javascript, CSN-Java, CSN-Go and 
CSN-PHP.

\begin{table}[h]
\def\arraystretch{1.0}
\setlength\tabcolsep{8pt} 
\begin{tabular}{@{}lllcccccc@{}}

\toprule
Model                            & \multicolumn{1}{l}{Method}             & \multicolumn{1}{c}{40}           
& \multicolumn{1}{c}{80}         & \multicolumn{1}{c}{120}               & \multicolumn{1}{c}{160}     
& \multicolumn{1}{c}{200}        \\ \midrule

\multirow{2}{*}{CodeT5p-220M}      
& InfoNCE                   & \multicolumn{1}{c}{18.8}                & \multicolumn{1}{c}{28.3}          
                            & \multicolumn{1}{c}{32.8}                & \multicolumn{1}{c}{35.3}          
                            & \multicolumn{1}{c}{35.5}                   
                            \\ \cmidrule(l){2-7} 
& URECA                     & \multicolumn{1}{c}{19.9(+1.1)}          & \multicolumn{1}{c}{28.4(+0.1)}          
                            & \multicolumn{1}{c}{35.3(+2.5)}          & \multicolumn{1}{c}{36.4(+1.1)}          
                            & \multicolumn{1}{c}{38.2(+2.7)}                  
                            \\ \midrule

\multirow{2}{*}{UniXCoder} 
& InfoNCE                   & \multicolumn{1}{c}{61.2}                & \multicolumn{1}{c}{61.4}          
                            & \multicolumn{1}{c}{63}                  & \multicolumn{1}{c}{62.9}                
                            & \multicolumn{1}{c}{63.8}                  
                            \\ \cmidrule(l){2-7} 
& URECA                     & \multicolumn{1}{c}{61.5(-0.3)}          & \multicolumn{1}{c}{61.4(+0)}          
                            & \multicolumn{1}{c}{63.4(+0.4)}          & \multicolumn{1}{c}{63.6(+0.7)}          
                            & \multicolumn{1}{c}{64.2(+04)}                   
                            \\ \midrule

\multirow{2}{*}{CoCoSoDA} 
& InfoNCE                   & \multicolumn{1}{c}{75}                  & \multicolumn{1}{c}{75.5}          
                            & \multicolumn{1}{c}{75.7}                & \multicolumn{1}{c}{76.0}          
                            & \multicolumn{1}{c}{76.4}                    
                            \\ \cmidrule(l){2-7} 
& URECA                     & \multicolumn{1}{c}{76.8(+1.8)}          & \multicolumn{1}{c}{77.8(+2.3)}          
                            & \multicolumn{1}{c}{78.8(+3.1)}          & \multicolumn{1}{c}{79.0(+3)}          
                            & \multicolumn{1}{c}{79.5(+3.1)}                   
                            \\ \bottomrule 
\end{tabular}
\caption{Results of Ruby across different number of few shot examples (MRR).}
\label{CSN_Ruby}
\end{table}

\begin{table}[h]
\def\arraystretch{1.0}
\setlength\tabcolsep{8pt} 
\begin{tabular}{@{}lllcccccc@{}}

\toprule
Model                            & \multicolumn{1}{l}{Method}              & \multicolumn{1}{c}{40}           
& \multicolumn{1}{c}{80}         & \multicolumn{1}{c}{120}                 & \multicolumn{1}{c}{160}     
& \multicolumn{1}{c}{200}        \\ \midrule

\multirow{2}{*}{CodeT5p-220M}      
& InfoNCE                   & \multicolumn{1}{c}{15.1}          & \multicolumn{1}{c}{21.6}          
                            & \multicolumn{1}{c}{24}          & \multicolumn{1}{c}{22}          
                            & \multicolumn{1}{c}{25.2}                   
                            \\ \cmidrule(l){2-7} 
& URECA                     & \multicolumn{1}{c}{14.7(-0.4)}          & \multicolumn{1}{c}{23.2(+1.7)}          
                            & \multicolumn{1}{c}{24(+0.9)}          & \multicolumn{1}{c}{23.2(+1.2)}          
                            & \multicolumn{1}{c}{26.4(+1)}                  
                            \\ \midrule

\multirow{2}{*}{UniXCoder} 
& InfoNCE                   & \multicolumn{1}{c}{48.9}              & \multicolumn{1}{c}{50.5}          
                            & \multicolumn{1}{c}{48.4}              & \multicolumn{1}{c}{51.2}          
                            & \multicolumn{1}{c}{51.4}                  
                            \\ \cmidrule(l){2-7} 
& URECA                     & \multicolumn{1}{c}{48.9(+0)}          & \multicolumn{1}{c}{51.1(+0.6)}          
                            & \multicolumn{1}{c}{51.7(+3.3)}          & \multicolumn{1}{c}{52.6(+2.4)}          
                            & \multicolumn{1}{c}{54(+2.6)}                   
                            \\ \midrule

\multirow{2}{*}{CoCoSoDA} 
& InfoNCE                   & \multicolumn{1}{c}{61.4}              & \multicolumn{1}{c}{62.5}          
                            & \multicolumn{1}{c}{63.4}              & \multicolumn{1}{c}{63.7}          
                            & \multicolumn{1}{c}{64.3}                    
                            \\ \cmidrule(l){2-7} 
& URECA                     & \multicolumn{1}{c}{64.4(+3)}          & \multicolumn{1}{c}{65.3(+2.8)}          
                            & \multicolumn{1}{c}{67(+3.6)}          & \multicolumn{1}{c}{68.4(+4.7)}          
                            & \multicolumn{1}{c}{68.8(+4.5)}                   
                            \\ \bottomrule 
\end{tabular}
\caption{Results of Javascript across different number of few shot examples (MRR).}
\label{CSN_Javascript}
\end{table}

\begin{table*}[h]
\def\arraystretch{1.0}
\setlength\tabcolsep{8pt} 
\begin{tabular}{@{}lllcccccc@{}}

\toprule
Model                            & \multicolumn{1}{l}{Method}              & \multicolumn{1}{c}{40}           
& \multicolumn{1}{c}{80}         & \multicolumn{1}{c}{120}               & \multicolumn{1}{c}{160}     
& \multicolumn{1}{c}{200}        \\ \midrule

\multirow{2}{*}{CodeT5p-220M}      
& InfoNCE                   & \multicolumn{1}{c}{3.1}               & \multicolumn{1}{c}{9.8}          
                            & \multicolumn{1}{c}{22}                & \multicolumn{1}{c}{22.6}          
                            & \multicolumn{1}{c}{26.9}                    
                            \\ \cmidrule(l){2-7} 
& URECA                     & \multicolumn{1}{c}{3.1(+0)}         & \multicolumn{1}{c}{11(+1.2)}          
                            & \multicolumn{1}{c}{20.2(-1.8)}        & \multicolumn{1}{c}{25.8(+3.2)}          
                            & \multicolumn{1}{c}{29.4(+2.5)}                
                            \\ \midrule

\multirow{2}{*}{UniXCoder} 
& InfoNCE                   & \multicolumn{1}{c}{52.6}             & \multicolumn{1}{c}{52.7}          
                            & \multicolumn{1}{c}{54.2}             & \multicolumn{1}{c}{55.1}             
                            & \multicolumn{1}{c}{55.5}                   
                            \\ \cmidrule(l){2-7} 
& URECA                     & \multicolumn{1}{c}{52.4(-0.2)}          & \multicolumn{1}{c}{54(+1.3)}          
                            & \multicolumn{1}{c}{55.4(+0.8)}          & \multicolumn{1}{c}{57.2(+2.1)}          
                            & \multicolumn{1}{c}{58.2(+2.7)}                  
                            \\ \midrule

\multirow{2}{*}{CoCoSoDA} 
& InfoNCE                   & \multicolumn{1}{c}{61.1}          & \multicolumn{1}{c}{61.7}          
                            & \multicolumn{1}{c}{64}          & \multicolumn{1}{c}{64.3}          
                            & \multicolumn{1}{c}{65}                    
                            \\ \cmidrule(l){2-7} 
& URECA                     & \multicolumn{1}{c}{63(+1.9)}        & \multicolumn{1}{c}{64.6(+2.9)}          
                            & \multicolumn{1}{c}{66.1(+2.1)}          & \multicolumn{1}{c}{68.1(+3.8)}          
                            & \multicolumn{1}{c}{68.4(+3.4)}                   
                            \\ \bottomrule 
\end{tabular}
\caption{Results of Java across different number of few shot examples (MRR).}
\label{CSN_Java}
\end{table*}

\begin{table*}[h]
\def\arraystretch{1.0}
\setlength\tabcolsep{8pt} 
\begin{tabular}{@{}lllcccccc@{}}

\toprule
Model                            & \multicolumn{1}{l}{Method}              & \multicolumn{1}{c}{40}           
& \multicolumn{1}{c}{80}         & \multicolumn{1}{c}{120}               & \multicolumn{1}{c}{160}     
& \multicolumn{1}{c}{200}        \\ \midrule

\multirow{2}{*}{CodeT5p-220M}      
& InfoNCE                   & \multicolumn{1}{c}{26.5}                & \multicolumn{1}{c}{43.1}          
                            & \multicolumn{1}{c}{42}                & \multicolumn{1}{c}{65.7}          
                            & \multicolumn{1}{c}{66.8}                  
                            \\ \cmidrule(l){2-7} 
& URECA                     & \multicolumn{1}{c}{28.8(+2.3)}          & \multicolumn{1}{c}{46.9(+3.8)}          
                            & \multicolumn{1}{c}{59.1(+17.1)}          & \multicolumn{1}{c}{66.9(+1.2)}          
                            & \multicolumn{1}{c}{68.3(+1.5)}                   
                            \\ \midrule

\multirow{2}{*}{UniXCoder} 
& InfoNCE                   & \multicolumn{1}{c}{70.5}               & \multicolumn{1}{c}{73.4}          
                            & \multicolumn{1}{c}{73.8}               & \multicolumn{1}{c}{80.2}          
                            & \multicolumn{1}{c}{80.1}                   
                            \\ \cmidrule(l){2-7} 
& URECA                     & \multicolumn{1}{c}{70.5(+0)}          & \multicolumn{1}{c}{74.8(+1.4)}          
                            & \multicolumn{1}{c}{74.6(+0.8)}          & \multicolumn{1}{c}{80.2(+0)}          
                            & \multicolumn{1}{c}{81.3(+1.2)}                  
                            \\ \midrule

\multirow{2}{*}{CoCoSoDA} 
& InfoNCE                   & \multicolumn{1}{c}{75.2}             & \multicolumn{1}{c}{77.3}          
                            & \multicolumn{1}{c}{79.1}             & \multicolumn{1}{c}{79.2}          
                            & \multicolumn{1}{c}{78.1}                    
                            \\ \cmidrule(l){2-7} 
& URECA                     & \multicolumn{1}{c}{82(+6.8)}          & \multicolumn{1}{c}{83.4(+6.1)}          
                            & \multicolumn{1}{c}{83.7(+4.6)}          & \multicolumn{1}{c}{83.3(+4.1)}          
                            & \multicolumn{1}{c}{83.4(+5.3)}                   
                            \\ \bottomrule 
\end{tabular}
\caption{Results of Go across different number of few shot examples (MRR).}
\label{CSN_Go}
\end{table*}

\begin{table*}[h]
\def\arraystretch{1.0}
\setlength\tabcolsep{8pt} 
\begin{tabular}{@{}lllcccccc@{}}

\toprule
Model                            & \multicolumn{1}{l}{Method}              & \multicolumn{1}{c}{40}           
& \multicolumn{1}{c}{80}         & \multicolumn{1}{c}{120}               & \multicolumn{1}{c}{160}     
& \multicolumn{1}{c}{200}        \\ \midrule

\multirow{2}{*}{CodeT5p-220M}      
& InfoNCE                   & \multicolumn{1}{c}{1.8}          & \multicolumn{1}{c}{11.4}          
                            & \multicolumn{1}{c}{15.8}          & \multicolumn{1}{c}{20.5}          
                            & \multicolumn{1}{c}{21.9}                  
                            \\ \cmidrule(l){2-7} 
& URECA                     & \multicolumn{1}{c}{2.6(+0.8)}          & \multicolumn{1}{c}{18(+6.6)}          
                            & \multicolumn{1}{c}{17.5(+1.7)}          & \multicolumn{1}{c}{21.9(+1.3)}          
                            & \multicolumn{1}{c}{23.4(+5.3)}                   
                            \\ \midrule

\multirow{2}{*}{UniXCoder} 
& InfoNCE                   & \multicolumn{1}{c}{40.4}          & \multicolumn{1}{c}{46.3}          
                            & \multicolumn{1}{c}{48.6}          & \multicolumn{1}{c}{48.4}          
                            & \multicolumn{1}{c}{50.4}                   
                            \\ \cmidrule(l){2-7} 
& URECA                     & \multicolumn{1}{c}{40.5(+0.1)}          & \multicolumn{1}{c}{44.8(-1.5)}          
                            & \multicolumn{1}{c}{47.6(-1.1)}          & \multicolumn{1}{c}{50.4(+2)}          
                            & \multicolumn{1}{c}{51.6(+1.2)}                  
                            \\ \midrule

\multirow{2}{*}{CoCoSoDA} 
& InfoNCE                   & \multicolumn{1}{c}{54.2}          & \multicolumn{1}{c}{55.9}          
                            & \multicolumn{1}{c}{55.9}          & \multicolumn{1}{c}{57.1}          
                            & \multicolumn{1}{c}{56.8}                    
                            \\ \cmidrule(l){2-7} 
& URECA                     & \multicolumn{1}{c}{57.6(+3.4)}          & \multicolumn{1}{c}{60.2(+5.3)}          
                            & \multicolumn{1}{c}{60.3(+5.4)}          & \multicolumn{1}{c}{61.2(+4.1)}          
                            & \multicolumn{1}{c}{60.3(+3.5)}                   
                            \\ \bottomrule 
\end{tabular}
\caption{Results of PHP across different number of few shot examples (MRR).}
\label{CSN_PHP}
\end{table*}

\newpage
\subsection{Thresholdly-Updatable Stationary Assumption for Dynamics}
\begin{table*}[h]
\def\arraystretch{1.0}
\setlength\tabcolsep{8pt} 
\begin{tabular}{@{}lllcccccc@{}}

\toprule
Model                            & \multicolumn{1}{l}{Dataset}              & \multicolumn{1}{c}{40}           
& \multicolumn{1}{c}{80}         & \multicolumn{1}{c}{120}               & \multicolumn{1}{c}{160}     
& \multicolumn{1}{c}{200}        \\ \midrule

\multirow{2}{*}{CSN-Python} 
& Stationary                & \multicolumn{1}{c}{63.2}          & \multicolumn{1}{c}{63.6}          
                            & \multicolumn{1}{c}{63}          & \multicolumn{1}{c}{63.7}          
                            & \multicolumn{1}{c}{64.7}                   
                            \\ \cmidrule(l){2-7} 
& URECA                     & \multicolumn{1}{c}{64.9(+1.7)}          & \multicolumn{1}{c}{65.8(+2.2)}          
                            & \multicolumn{1}{c}{66.6(+3.6)}          & \multicolumn{1}{c}{67.2(+3.5)}          
                            & \multicolumn{1}{c}{67.7(+3)}                  
                            \\ \midrule

\multirow{2}{*}{CoSQA} 
& Stationary                & \multicolumn{1}{c}{48.6}          & \multicolumn{1}{c}{49.9}          
                            & \multicolumn{1}{c}{51.6}          & \multicolumn{1}{c}{53.9}          
                            & \multicolumn{1}{c}{54.4}                    
                            \\ \cmidrule(l){2-7} 
& URECA                     & \multicolumn{1}{c}{52(+4)}          & \multicolumn{1}{c}{54.6(+4.7)}          
                            & \multicolumn{1}{c}{54.6(+3)}          & \multicolumn{1}{c}{56.3(+2.2)}          
                            & \multicolumn{1}{c}{55.6(+1.2)}                   
                            \\ \bottomrule 
\end{tabular}
\caption{CoCoSoDA Results of Stationary Assumption and URECA (Thresholdly Updatable Stationary Assumption) 
across different number of few shot examples (MRR).}
\label{Stationary Assumption}
\end{table*}

\subsection{Structure of Disentangled Representation}
The performance improvement of GraphCodeBERT in CSN-Python and CoSQA is not as significant 
as UniXcoder and CodeT5+ in Table 12.
This is because GraphCodeBERT incorporates dataflow into the input based on graph-guided attention,
allowing the model to learn the inherent semantic structure of the code.
Therefore, GraphCodeBERT has already sufficiently learned the inherent semantic structure
that others might not have fully captured.
However, for CoNaLa-docprompting dataset as dataset for pure nl dataset, the effects of GraphCodeBERT 
are limited since CoNaLa-docprompting requires model to retrieve nl-document rather than code, 
which makes dataflow and graph-guided-masked-attention impossible to apply. 

On CoNaLa (used validation split as reported in the original paper (\cite{PengchenBEBG18})), 
URECA achieved a recall@10 of 55.4, outperforming InfoNCE’s 49.1 by +6.3, 
since GraphcodeBERT's methodlogies require sufficient pre-training for the model to become familiar with dataflow and graph-guided attention
which is only usable for code domain. 
This suggests that URECA is more broadly effective across modalities than GraphCodeBERT.
In addition, GraphCodeBERT needs 5 to 6 times more seconds since graph-guided attention is always calculated 
for each iteration.
In contrast, URECA can be more generally applied since it only requires fine-tuning 
without these additional steps with a little time lag in reference to Figure 4.

\begin{table}[h]
\centering
\def\arraystretch{0.8}
\setlength\tabcolsep{8pt} 
\begin{tabular}{@{}lccc@{}}

\toprule
\multicolumn{1}{l}{Method}              & \multicolumn{1}{c}{CSN}           
& \multicolumn{1}{c}{CoSQA}          & \multicolumn{1}{c}{CoNaLa(R@10)}                \\ \midrule

InfoNCE                     & \multicolumn{1}{c}{49.4}          & \multicolumn{1}{c}{8.8}          
                            & \multicolumn{1}{c}{49.1}                          
                            \\ \midrule 
URECA                       & \multicolumn{1}{c}{49.7(+0.3)}          & \multicolumn{1}{c}{9.1(+0.3)}          
                            & \multicolumn{1}{c}{55.4(+6.3)}                             
                            \\ \bottomrule 
\end{tabular}
\caption{Results of GraphCodeBERT}
\label{overall}
\end{table}

\subsection{Implementation Details}
We fix the query and code length to 128 for UniXCoder, CodeT5p-220M, and CoCoSoDA in few-shot learning. 
For GraphCodeBERT, we set the query and data flow length to 64 and the code length to 256.
The batch size is selected from 12 and 24, choosing the value that yields the best performance. 
The learning rate is set to $2e-5$ for CoCoSoDA and $1e-5$ for UniXCoder, CodeT5p-220M, and GraphCodeBERT.
In few-shot scenario, all experiments are conducted on a GeForce RTX 3090 GPU, 
and each experiment is repeated three times using different random seeds for 40, 80, 120, 160, and 200 examples.
For CoNaLa-docprompting, we fix the maximum query and code length to 64 exclusively for URECA. 
In this case, the experiments are conducted on an A6000 GPU.
Our codes is based on SoftInfoNCE (\cite{HaochenXAC23}), and it is available at https://github.com/github\_id/ureca.

\end{document}